\documentclass[preprint,12pt]{elsarticle}

\usepackage{amssymb}
\usepackage{amsmath}
\usepackage{algorithm}
\usepackage{algorithmic}
\usepackage{graphicx}
\usepackage{rotating}
\usepackage{hyperref}

\usepackage{tikz}
\usetikzlibrary{arrows.meta, positioning, shapes, decorations.pathreplacing, calc}

\usepackage{booktabs}
\usepackage{siunitx}
\usepackage[percent]{overpic}
\usepackage[utf8]{inputenc}
\usepackage{siunitx}
\usepackage{graphicx}

\usepackage{subcaption}

\definecolor{myGreen}{rgb}{0.4501, 0.6784,0.2745}

\definecolor{myblue}{rgb}{0.705882, 0.78039, 0.90588}

\definecolor{mywhite}{rgb}{1,1,1}
\definecolor{blue2}{rgb}{0.7098,0.78039,0.913725}

\definecolor{blueshape}{rgb}{0.670589,0.74902,0.901961}

\definecolor{greenframe}{rgb}{0.77255,1,0.77255}

\definecolor{greenabstract}{rgb}{0.55691,0.851,0.451}

\definecolor{blueabstract}{rgb}{0.27451,0.6941,0.88235}

\journal{}

\begin{document}

\begin{frontmatter}




\title{A Reward-Directed Diffusion Framework for Generative Design Optimization}

\author[label1]{Hadi Keramati\corref{cor1}}

\author[label1]{Patrick Kirchen}\
\author[label1]{Mohammed Hannan}\
\author[label1]{Rajeev K. Jaiman}\

\cortext[cor1]{Corresponding author: hadi.keramati@ubc.ca}

\affiliation[label1]{organization={Department of Mechanical Engineering},
            addressline={University of British Columbia}, 
            city={Vancouver}, 
            state={BC},
            country={Canada}}

\begin{abstract}

This study presents a generative optimization framework that builds on a fine-tuned diffusion model and reward-directed sampling to generate high-performance engineering designs.
The framework adopts a parametric representation of the design geometry and produces new parameter sets corresponding to designs with enhanced performance metrics.
A key advantage of the reward-directed approach is its suitability for scenarios in which performance metrics rely on costly engineering simulations or surrogate models (e.g. graph‑based, ensemble models, or tree‑based) are non‑differentiable or prohibitively expensive to differentiate. This work introduces the iterative use of a soft value function within a Markov decision process framework to achieve reward-guided decoding in the diffusion model. By incorporating soft-value guidance during both the training and inference phases, the proposed approach reduces computational and memory costs to achieve high-reward designs, even beyond the training data.
Empirical results indicate that this iterative reward-directed method substantially improves the diffusion model's ability to generate samples with reduced resistance in 3D ship hull design and enhanced hydrodynamic performance in 2D airfoil design tasks. The proposed framework generates samples that extend beyond the training data distribution, resulting in a greater 25\% reduction in resistance for ship design and over 10\% improvement in the lift-to-drag ratio for the 2D airfoil design.
Successful integration of this model into the engineering design life cycle can enhance both designer productivity and overall design performance.

\end{abstract}


\begin{graphicalabstract}
 \centering
\setlength{\fboxsep}{0.05pt} 
 \begin{overpic}[width=1\linewidth,grid=false
]{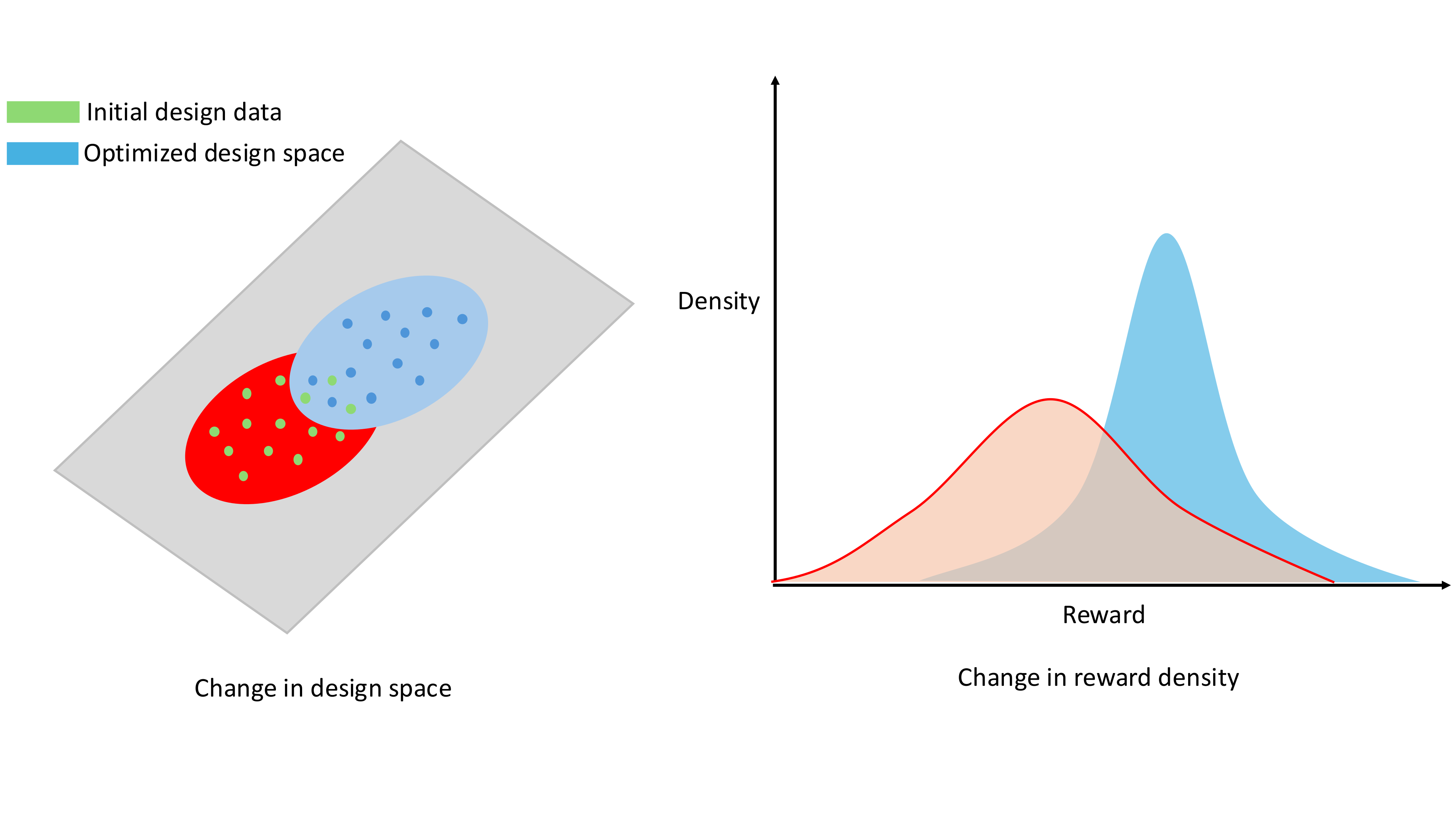}
\put(6,7.6){\colorbox{white}{\strut \text{Change in design samples}}}

\put(57.5,8){\colorbox{white}{\strut \text{Change in reward density}}}

\put(70,12.5){\colorbox{white}{\strut \text{Reward}}}

\put(44,35){\colorbox{white}{\scriptsize\strut  \hspace{32pt}}}

\put(42.7,32){\colorbox{white}{\strut \text{Density}}}


\put(0,47.95){\colorbox{white}{\scriptsize\strut  \hspace{100pt}}}

\put(-14,48.15){\colorbox{greenabstract}{\textcolor{greenabstract}{\rule{18pt}{5pt}}}}

\put(-8.2,47.95){\strut \text{Initial design data}}




\put(0,45.6){\colorbox{white}{\scriptsize\strut  \hspace{95pt}}}

\put(-14,45.1){\colorbox{blueabstract}{\textcolor{blueabstract}{\rule{18pt}{5pt}}}}

\put(-9.5,44.3){\strut \text{ Optimized design sample}}


\end{overpic}
\end{graphicalabstract}


\end{frontmatter}



\section{Introduction}
\label{sec1}
Engineering design is a complex process that involves the application of scientific computing, physics, and creative thinking to develop solutions that meet specific needs or solve particular problems \cite{elmaraghy2012complexity,gaspar_2012,misra2020design}. 
The design process is inherently iterative and requires a deep understanding of both the problem space and the potential solutions. Common practice in engineering design involves teams of designers and simulation engineers collaborating on theoretical and numerical aspects to produce feasible solutions for specific applications such as naval architecture, automotive engineering, and aircraft design. For example, the primary goal of ship designers is to design a hull shape that maximizes cargo carrying capacity while minimizing resistance through the water for a given ship displacement \cite{misra2020design,zhang2019research}. If we break down the ship optimization process, it is essentially a constrained optimization in which the designers consider hard or soft constraints on design variables. This process relies on the designer's experience and their perception of propulsion, which is extremely nonlinear with respect to the shape of the hull and the flow characteristics \cite{zhang2019research}. This is often a time-consuming and costly process with several limitations.
Furthermore, using existing datasets to optimize engineering design presents significant challenges for practical applications.  The objective values for optimizing the engineering design are obtained using numerical simulations, which are not differentiated with respect to the design variables \cite{kumar2020model}.

In recent years, the integration of data-driven approaches has significantly transformed the landscape of engineering design, allowing more efficient and innovative solutions \cite{brunton2022data,regenwetter2022deep}. Surrogate modeling has been used in the engineering community for a long time starting with Gaussian process regression known as the Kriging method, and later artificial neural networks were adopted \cite{jouhaud2006kriging,park2006application,kawai2014kriging,alizadehdakhel2009cfd,guo2016convolutional}. These surrogate models, combined with data-driven methods that use existing experimental data, can generate full-scale digital twins for both design and operational optimization \cite{fonseca2022standards,bronson2024challenges}. Physics-informed neural networks and learning operators were introduced to accelerate physics simulation and speed up the engineering design process \cite{raissi2019physics,li2020fourier}. For example, a physics-informed convolutional neural network (CNN) is used to predict temperature fields in heat source layouts without labeled data \cite{zhao2023physics}. The model leverages physical knowledge, specifically the heat conduction equation, to train the CNN, aiming to minimize violations of physical laws, rather than relying on large datasets. Bokil et al. \cite{bokil2024physics} presented a physics-guided CNN to predict flow characteristics in heat exchangers for electrified aircraft to reduce the computational cost of simulations to perform design optimization while maintaining the accuracy of the simulations. 

\begin{figure}[h!]
    \centering
    \includegraphics[width=0.95\textwidth]{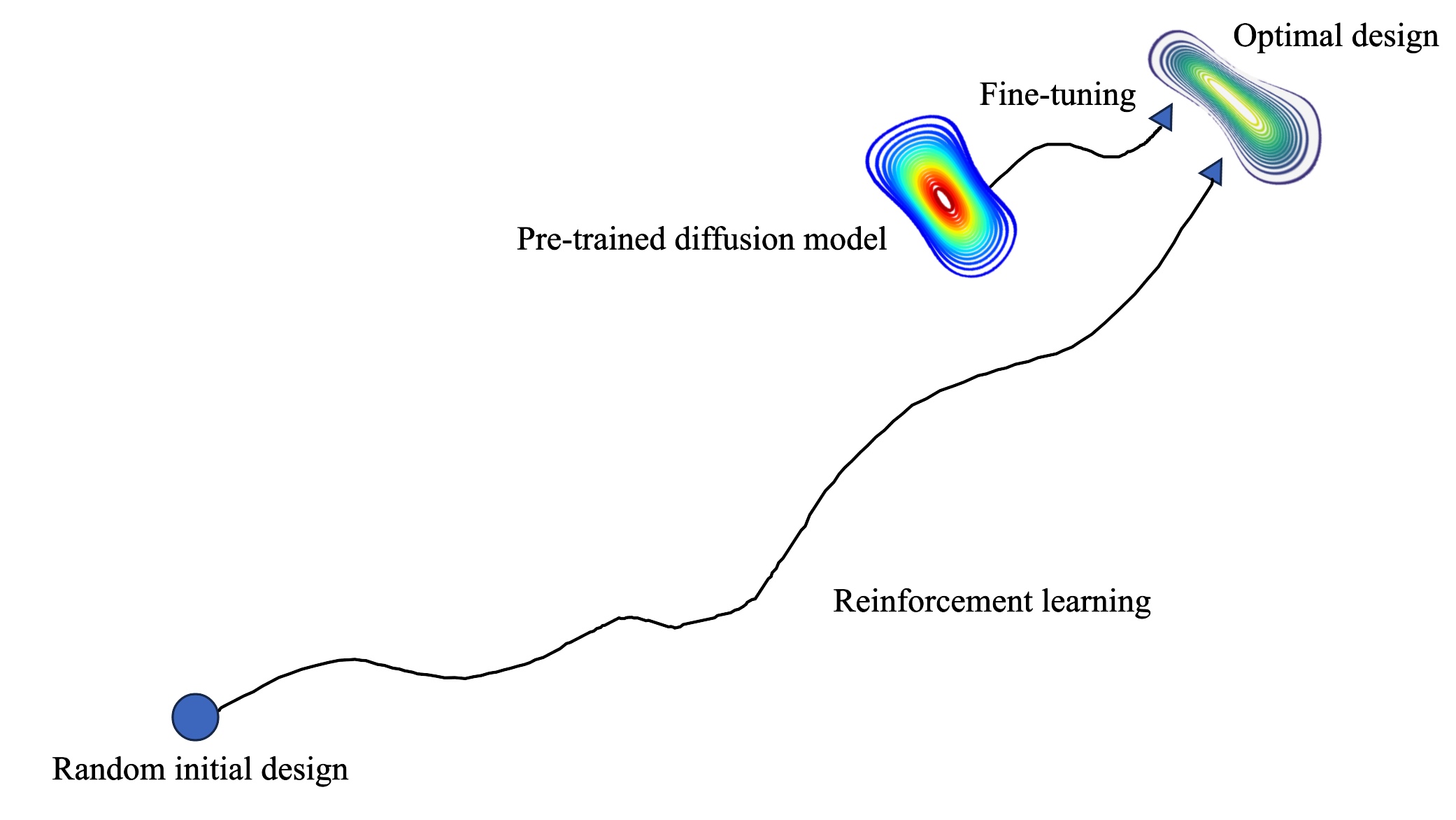} 
    \caption{A simple illustration of design optimization trajectories for reinforcement learning and fine-tuned diffusion model. The background color map and contour lines represent the learned generative distribution.} 
    \label{rl-ddp} 
\end{figure}

Generative models for engineering design are based on either online simulation or existing datasets. Variational autoencoders (VAE), generative adversarial networks (GAN), and diffusion models are popular generative design frameworks that require existing data sets \cite{lew2021encoding, chen2019aerodynamic}. Reinforcement learning (RL) and adjoint-based topology optimization, on the other hand, require online setup to simulate the physics of engineering problems \cite{keramati2022generative, giannakoglou2008adjoint}. Treating design optimization as a reinforcement learning problem is computationally expensive and unstable, particularly in high‑dimensional spaces, because policy optimization begins from a random design and must navigate its way to an optimal solution. To address this, we employ reward‑based fine‑tuning to guide the model toward higher rewards rather than initiating policy optimization from scratch. Figure \ref{rl-ddp} illustrates the difference between pure reinforcement learning for design optimization and the fine-tuning approach presented in this paper. However, the process to reach the optimal design in this study involves two steps applied to a pre-trained model, rather than a single-step approach.
In reinforcement learning, the optimization trajectory begins from a random initial design and gradually improves through trial-and-error interactions with the reward function, often requiring many costly evaluations and suffering from unstable convergence in high-dimensional spaces. 
In contrast, the reward-directed diffusion model performs sampling from a pre-trained distribution, which is gradually shifted toward regions of higher reward via the fine-tuned diffusion model.

All generative design methods require data, and data curation poses a significant challenge in engineering applications. Several studies have been conducted to reduce the size of datasets required to train surrogate models with high accuracy \cite{baque2018geodesic, guo2016convolutional}. Baque et al.~\cite{baque2018geodesic} employed Geodesic Convolutional Neural Networks to decrease the size of the training set while maintaining the accuracy of the regression. The parameterization of engineering designs to perform modern machine learning frameworks for regression, such as Automated Machine Learning (AutoML), has become increasingly popular \cite{regenwetter2023framed}. This popularity is due to more convenient data processing and lower memory requirements compared to handling 3D data, such as meshes and other mathematical geometric representations \cite{wang2003level, da2018evolutionary, wang2004structural}.

Convolutional neural networks triggered more geometry components in CFD prediction that could be used for shape optimization \cite{guo2016convolutional}. Other operators such as geodesic convolutional neural networks \cite{baque2018geodesic} were introduced that can be used for shape optimization purposes. Some of the newly developed algorithms, such as physics-informed neural networks \cite{raissi2019physics}, and neural operators, have limited applications in shape optimization due to the complexity of flow and boundary conditions in 3D space \cite{azizzadenesheli2024neural,shukla2024deep,shukla2023deep}. Graph neural network (GNN) surrogate models, however, showed strong potential for complex geometry and physics \cite{suk2021equivariant,horie2022physics,shao2023pignn}. In order to more precisely consider the boundary conditions and physics simulation in the shape optimization, Hadizadeh et al. \cite{hadizadeh2024graph} used a graph neural network to predict the flow field around various airfoil shapes. This surrogate model was used along with multi-objective optimization for fluid-acoustic shape optimization applications. 

Although surrogate models reduce the optimization cost, traditional gradient-based optimization algorithms are still limited in design applications. Engineering design optimization is a non-linear problem, and there is no clear expression between the design variables and performance of the design that can be used for derivation. Even with such expressions, gradient-based optimizations trap in local optima that are far from optimized design. Black-box optimization algorithms such as genetic algorithm, particle swarm optimization, and recently reinforcement learning show strong capabilities to find global optima, but suffer from computational cost in the early stage of search \cite{keramati2022deep}. 
Therefore, data-driven approaches that use dimensionality reduction and offline datasets for optimization and automated design have gained attention \cite{chen2019aerodynamic}. Generative adversarial networks showed early success in design optimization tasks \cite{chen2021mo}, but are difficult to train. This difficulty becomes highlighted when design constraints are considered in a form of conditional GAN \cite{mirza2014conditional}. Variational autoencoders, which are a neural network architecture of the encoder-decoder to extract lower-dimensional latent variables from input data, have also been used to generate designs with performance higher than the original data set \cite{kingma2013auto,zhang20193d}. Both GANS and VAE require that the objective function be differentiable from the design variables to perform gradient-based optimization \cite{chen2021mo}. 

Another class of generative models called diffusion models works by gradually adding noise to the data over a sequence and then training a model to learn the denoising reverse process step by step to recover the data distribution to generate new samples that resemble the initial data \cite{sohl2015deep,ho2020denoising,kotelnikov2023tabddpm}. Popular diffusion models are the Denoising Diffusion Probabilistic Model (DDPM) and score‑based generative modeling or diffusion models \cite{ho2020denoising, song2021scorebasedgenerativemodelingstochastic}. DDPMs are based on a Markov chain that adds noise according to a discrete schedule and learns the reverse denoising process by optimizing a variational objective \cite{ho2020denoising}. However, score-based models learn the score at continuous noise levels in the form of a stochastic differential equation \cite{song2021scorebasedgenerativemodelingstochastic}. 
Using guidance to impose conditions on the denoising process of the diffusion model to manipulate the generated samples became popular for a variety of applications \cite{ho2022classifier,song2020score,dhariwal2021diffusion}. 
There are several methods to use diffusion models as maximization frameworks \cite{li2024diffusion,krishnamoorthy2023diffusion}. Conditional diffusion models that are based on classifier-free guidance \cite{ho2022classifierfreediffusionguidance} are used to generate high performance designs \cite{krishnamoorthy2023diffusion}.  The methods based on classifier-free models can generate high-reward designs that are inside the pre-trained distribution, but the performance of generating high-performance design structures is limited to the training data distribution \cite{krishnamoorthy2023diffusion}. For the purpose of design optimization, the goal is to achieve a performance beyond the training data. 

Conditional models are also used for optimization in engineering design \cite{bagazinski2024c}. The results for the design of generative ships show that it can be interpolation or extrapolation, but the framework is based on the gradient of the classifier and the performance regressor for optimization \cite{dhariwal2021diffusionmodelsbeatgans,bagazinski2023shipgen}. This requires the objective function for the optimization to be differentiable. For many engineering and scientific applications, this is not practical. The objective function in engineering and scientific application requires high-fidelity simulation or multi-fidelity methods \cite{sajjadinia2022multi,arie2017air}. In case of training a proxy for the objective function to use the gradient-based guidance, the choice of the proxy will be limited to differentiable models such as neural networks. Highly effective and efficient models, such as tree-based algorithms, ensemble methods \cite{ganaie2022ensemble}, as well as graph neural networks, are non-differentiable or computationally expensive and complex to perform differentiation during optimization \cite{wen2023spectral,hadizadeh2024graph}.  

In this study, we introduce a generative design optimization framework that does not require computing gradients with respect to design variables, making it practical for problems in which performance evaluation depends on high‑fidelity or multi‑fidelity simulations. This framework is designed to facilitate the use of non-differentiable machine learning models such as tree-based and ensemble models, and models with computationally expensive differentiation such as GNNs. We first parameterize the design, then pre-train a Denoising Diffusion Probabilistic Model to generate designs that resemble the training data. We use a unique guidance technique to distribute the computational and memory complexity of reward maximization to a training phase and inference time. We fine-tune the pre-trained DDPM using reward-weighted maximum likelihood estimation to tune the model parameters to generate designs from a distribution that is close to the target optimal policy. We then apply reward-based (value-based) sampling during inference to provide further alignment with optimized reward so that the generated samples are within maximum performance distribution. This framework is uniquely designed for reward functions that are based on physics-based simulation, physics-based surrogate models, and non-differentiable reward models. This method is suitable for scientific and engineering reward models. We perform experiments on 2D (airfoil) and 3D (ship hull) design problems to demonstrate the effectiveness of the proposed framework.

\section{Background and Preliminaries}
\label{sec:headings}
Before we present our proposed framework, we first provide a brief overview of diffusion models and their mathematical foundations. 
Diffusion models are a class of generative models that produce new data that are similar to the patterns found in their training sets. Essentially, they operate by gradually blurring the original data with a sequence of added Gaussian noise, then learning to reverse this process to reconstruct the data \cite{sohl2015deep}. This section briefly explains the DDPM to build a foundation for describing reward-based fine-tuning and sampling in the next section. DDPM is a latent variable model in which a forward diffusion process gradually adds noise to data of the same dimensionality \cite{ho2020denoising}. A learned reverse process, modeled as a Markov chain, iteratively removes the noise to reconstruct high-quality samples.

\subsection{Forward Diffusion Process}
\label{subsec1}
In the DDPM forward process, at each time step \( t \in \{1, 2, \dots, T\} \), where \( T \) is the final timestep in the forward process, Gaussian noise with variance schedule \( \beta_t \) is added to the input data \( {x}_0 \) at each step of the fixed Markov chain. The transition distribution \( q\) from step \( t - 1 \) to step \( t \) is formulated as follows:
\begin{equation}
q({x}_t \mid {x}_{t-1}) = \mathcal{N}({x}_t; \alpha_t {x}_{t-1}, \beta_t I)
\label{eq1}
\end{equation}
where $\alpha_t = 1 - \beta_t$, and $ I $ is identity matrix. The distribution at timestep \( t \) is defined as follows 
\begin{equation}
   {x}_t = \sqrt{\bar{\alpha}_t} \, {x}_0 + \sqrt{1 - \bar{\alpha}_t} \, \epsilon, \quad \epsilon\sim \mathcal{N}(0, I)
    \label{eq2}
\end{equation}
where $\bar{\alpha}_t = \prod_{s=1}^{t} \alpha_s$ \cite{ho2020denoising}. 


\subsection{Reverse Denoising Process}
\label{subsec2}
At timestep \( T \), the latent \( {x}_T \) is close to the isotropic Gaussian distribution. The reverse process chain is designed to recover \( {x}_0 \) from \( {x}_T \) by estimating the noise added at each timestep.  This estimation is performed by a neural network $p_\theta$, on the mean and variance assuming the noise is Gaussian similar to the forward process,
\begin{equation}
    p_\theta({x}_{t-1} \mid {x}_t) = \mathcal{N}({x}_{t-1} \mid \mu_\theta({x}_t, t), \beta_t I ),
    \label{eq3}
\end{equation}
 where $ \mu_\theta({x}_t, t)$ is a learnable estimator for the mean of ${x}_{t-1}$ that only depends on ${x}_t$ at timestep $t$
\begin{equation}
    {x}_{t-1} = \frac{1}{\sqrt{\alpha_t}} \left( {x}_t - \frac{\beta_t}{\sqrt{1 - \bar{\alpha}_t}} \, \epsilon_\theta({x}_t, t) \right) + \sigma_t \, z, \quad z \sim \mathcal{N}(0, I).
    \label{eq4}
\end{equation}
Here, \( \epsilon_\theta({x}_t, t) \) is the estimation of noise from the neural network and \( \sigma_t \) is the variance of the added noise. This process proceeds from \(t = T\) down to \(t = 0\), and sampling from \(x_{0}\) yields new data that resemble the original data \cite{ho2020denoising}. 

\subsection{DDPM Training Objective}
\label{subsec3}
Training the DDPM is based on maximum likelihood estimation (MLE), which involves maximizing the data log-likelihood by optimizing a variational bound on the negative log-likelihood:
\begin{equation}
L_{\mathrm{MLE}}
\;=\;
-\mathbb{E}_{x_0}\bigl[\log p_{\theta}(x_{0})\bigr]
\;\le\;
\mathbb{E}_{q}\!\Bigl[
-\log \frac{p_{\theta}(x_{0:T})}{q(x_{1:T}\!\mid\!x_{0})}
\Bigr].
\label{eq:ddpm_elbo}
\end{equation}
In practice, this bound is optimized by minimizing the simplified
noise‑matching loss. In simple terms, DDPM is trained by minimizing the loss term defined by the distance between the predicted noise and the true noise \cite{ho2020denoising}:
\begin{equation}
\label{lddpm}
    \mathcal{L}_{\text{DDPM}} = \mathbb{E}_{{x}_0, \epsilon, t} \left[ \left\| \epsilon - \epsilon_\theta({x}_t, t) \right\|^2 \right].
\end{equation}
In a scenario where we have a pre-trained diffusion model, there are several methods such as classifier guidance that utilize this pre-trained model for conditional generation. In classifier guidance, the pre-trained model remains frozen, with no fine-tuning applied, and the method functions solely during inference by steering sample generation through gradient updates.
While diffusion models have proven powerful for generative modeling, applying them to engineering design optimization presents unique challenges, especially when rewards are derived from non-differentiable or simulation-based evaluations. In the next section, we introduce a reward-directed formulation for both the training and inference processes to bias the model toward high-performance designs without requiring gradients of the reward function with respect to the design parameters.

\section{Diffusion-Based Generative Optimization Framework}
This section explains our proposed methodology for iteratively using a soft value function for reward guidance using a fine-tuned diffusion model and reward-based importance sampling as an optimization method for shape optimization. First, we modify the objective of the diffusion model to incorporate reward values during fine-tuning of a pre-trained model. We use a reward-weighted maximum likelihood estimate to assign a higher likelihood to designs with higher rewards and guide the model further toward highly rewarded samples \cite{norouzi2016reward}. This distinguishes our approach from regression-based or classifier guidance because there is no need for a differentiable objective, which is extremely important for engineering applications. Our proposed method reformulates the design optimization problem into a framework with sequences of reward-guided components. The formulation is based on entropy-regularized Markov Decision Processes (MDPs) and importance sampling. As a result, it eliminates the requirement for differentiable reward functions and provides a key advantage in many engineering applications.

\vspace{10pt}
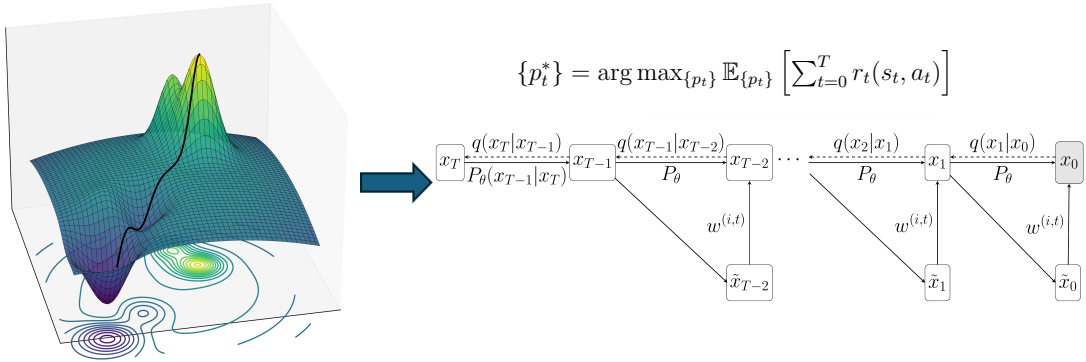
\begin{figure}[h]
   
    \centering
\setlength{\fboxsep}{0.6pt}
  \begin{overpic}[width= \textwidth,grid=false]{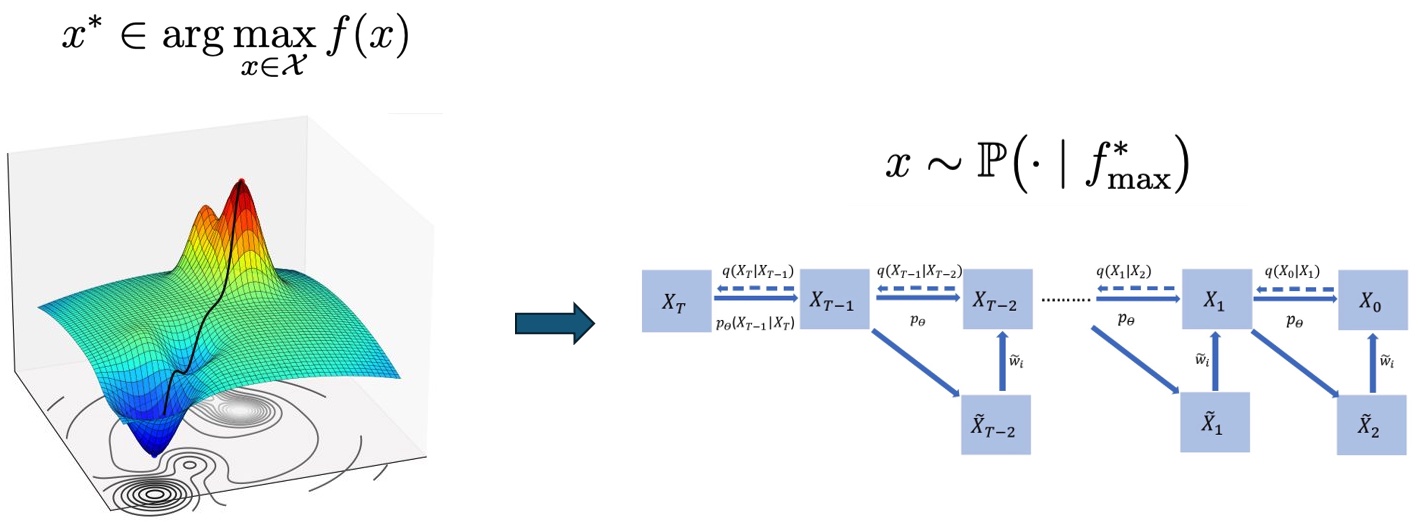}

  \put(-17,-4){\includegraphics[width=0.6\textwidth,]{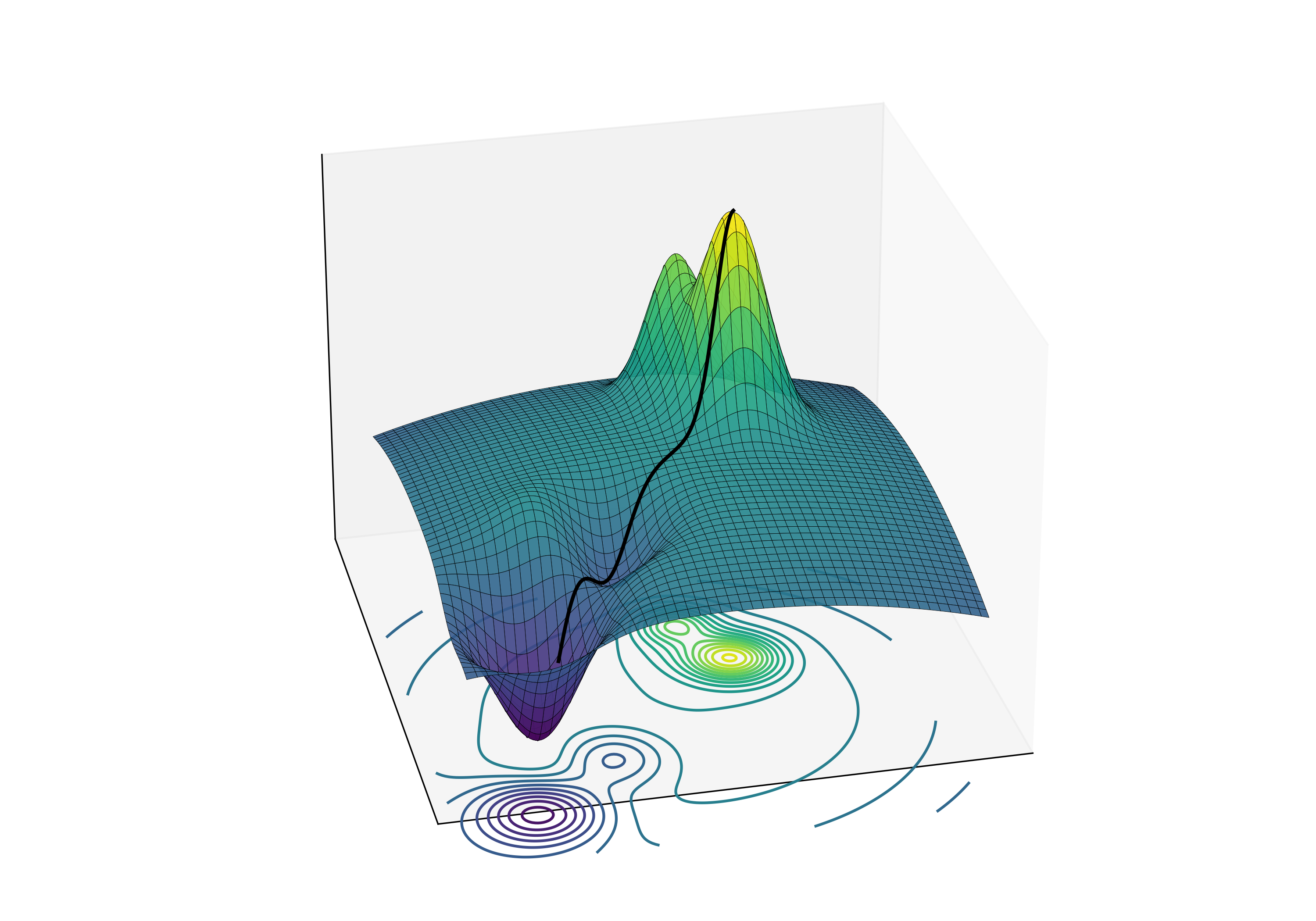}}

  \put(39.2,4){\includegraphics[width=0.63\textwidth]{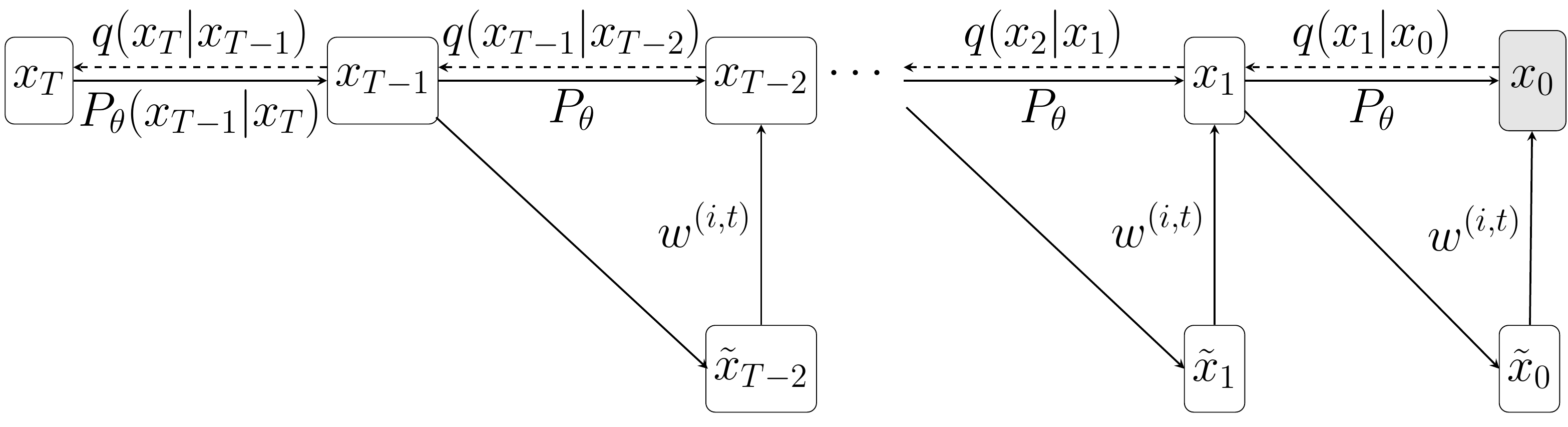}}

  \put(32,14){\includegraphics[width=0.07\textwidth]{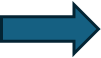}}

\put(-0.5,38){\colorbox{white}{\makebox[80 pt][l]{\rule{0pt}{5pt}}}}

\put(4.3,37){\colorbox{white}{\scalebox{0.8}{$\displaystyle x^* \in \arg \max_{x \in \mathcal{X}} f(x)$}}}

\put(62,23){\colorbox{white}{\makebox[85pt][l]{\rule{0pt}{30pt}}}}

\put(47,26){\colorbox{white}{\scalebox{0.8}{$
\{p_t^*\} = \arg \max_{\{p_t\}} \mathbb{E}_{\{p_t\}} \left[ \sum_{t=0}^{T} r_t(s_t, a_t) \right]$}}}


  \end{overpic}
    
    
    \caption{Conceptual schematic of the generative design optimization framework for shape optimization using reward-directed sampling from optimal policy } 
    \label{optim}
\end{figure}

\begin{figure}[h]
    \centering
    \setlength{\fboxsep}{0.6pt}
  \begin{overpic}[width= \textwidth,grid=false]{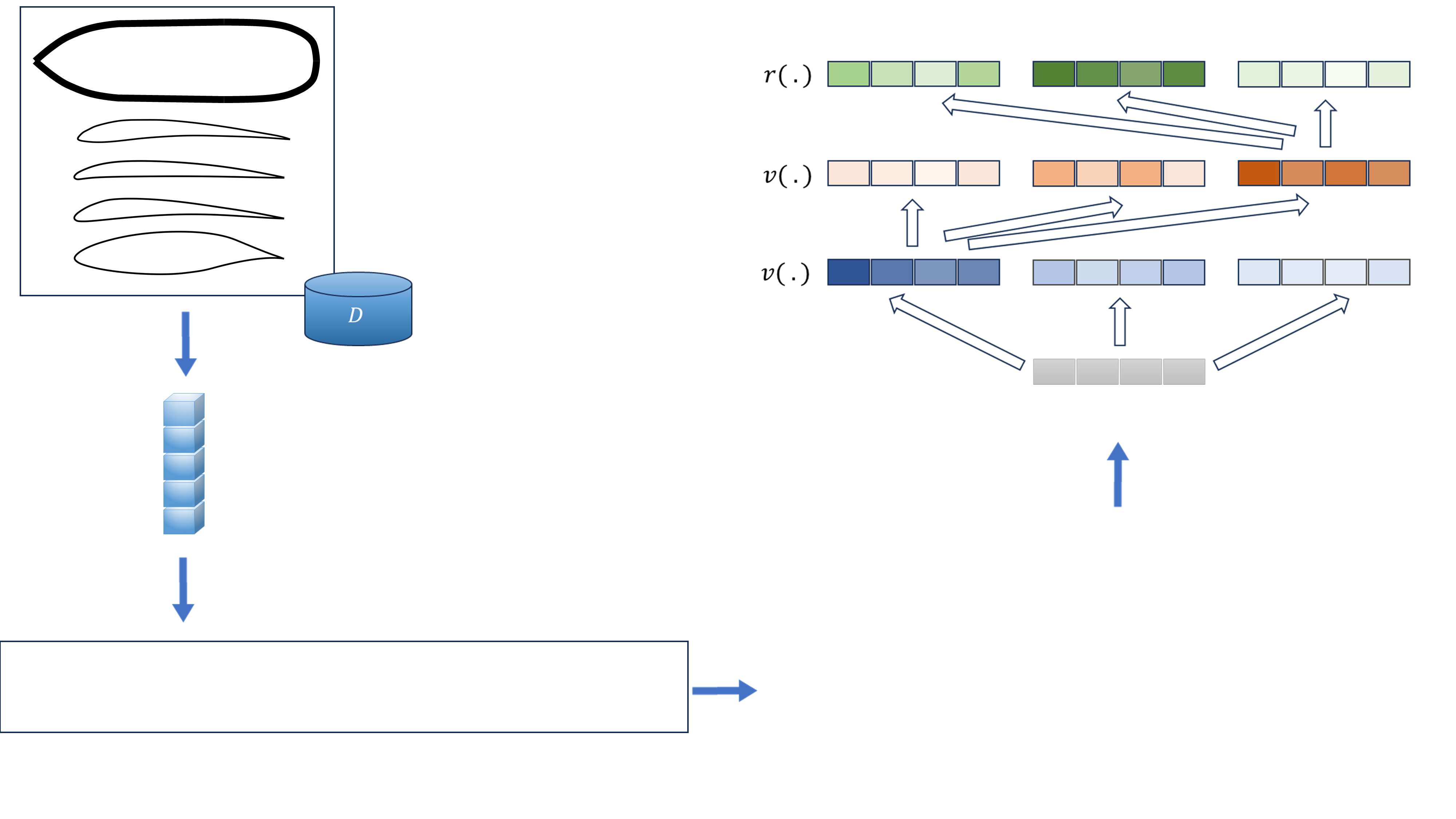}



\put(-7.5,24){\colorbox{white}{\fontsize{8pt}{2.4pt}\selectfont\text{Design vector data}}}

\put(51.88,5){\includegraphics[width=0.49\textwidth,]{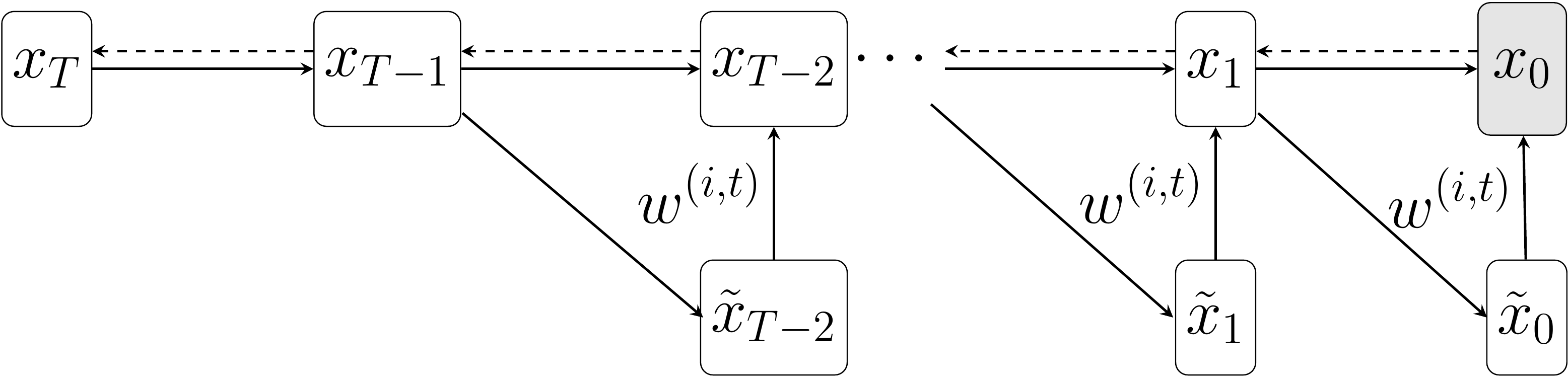}}
\put(0.17,6.5){\includegraphics[width=0.47\textwidth,]{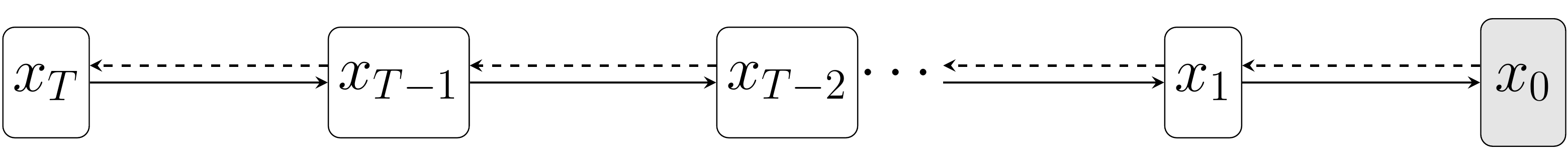}}

\put(18,1.5){\colorbox{white}{\fontsize{8pt}{2.4pt}\selectfont\text{Pre-training }}}

\put(77,1.1){\colorbox{white}{\scriptsize\strut  \vspace{10pt}\hspace{40pt}}}

\put(72,1.5){\colorbox{white}{\fontsize{8pt}{2.4pt}\selectfont\text{Fine-tuning}}}

\put(36,41){\colorbox{white}{\hspace{2pt}{\fontsize{8pt}{2.4pt}\selectfont\text{Value-weighted}}}}
\put(40,39){\fontsize{8pt}{2.4pt}\selectfont\text{sampling}}

\put(63,54){\colorbox{white}{\hspace{2pt}{\fontsize{7.5pt}{2.4pt}\selectfont\text{Samples from optimal policy}\hspace{4pt}}}}


\put(52,37){\colorbox{white}{\scriptsize\strut  \vspace{12pt}\hspace{13pt}}}
\put(52,37){\scriptsize $v(\cdot)$}

\put(52.5,43.7){\colorbox{white}{\scriptsize\strut  \vspace{11pt}\hspace{12pt}}}
\put(52,44){\scriptsize $v(\cdot)$}

\put(52.5,50.5){\colorbox{white}{\scriptsize\strut  \vspace{11pt}\hspace{12pt}}}
\put(52,50.5){\scriptsize $r(\cdot)$}

  \end{overpic}
    \caption{Visual workflow of the reward-directed generative optimization framework. The framework shows how the engineering data is used for pre-training, fine-tuning, and reward-based importance sampling.} 
    \label{iterativeschem} 
\end{figure}
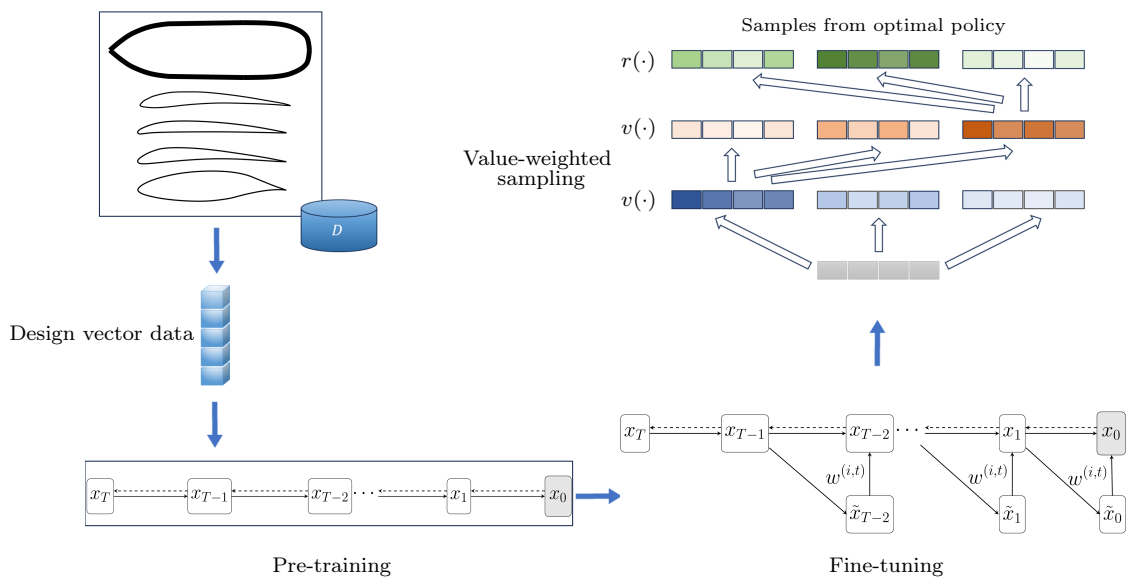

Fig. \ref{optim} illustrates how the optimization problem is reformulated as sampling from the optimal policy of a fine‑tuned model, while Fig. \ref{iterativeschem} outlines the overall framework and the iterative data flow for generating new high‑reward samples. We first perform pre-training without guidance using the loss function presented in Equation \ref{lddpm}. Then, we fine-tune the model using reward-weighted MLE, and sample the fine-tuned model using reward-based importance sampling~\cite{norouzi2016reward,peters2010relative}. Therefore, reward guidance is used during both training and inference.  To summarize, there are four key steps for generating high-performance designs namely: (i) parameterize the design using a parametric or vector representation, (ii) pretrain a diffusion model to generate designs similar to the original data, (iii) fine-tune the diffusion model using reward-weighted MLE, and (iv) optimize inference toward higher performance designs using reward-based importance sampling. 

\subsection{Markov Decision Process}
%
In the context of DDPM, we define the Markov decision process as follows:
\begin{equation}
\bigl\{\,\mathcal{S},\,\mathcal{A},\,\{P_t\}_{t=0}^T,\,
\{r_t\}_{t=0}^T\bigr\}    
\label{eq6}
\end{equation}
where $\mathcal{S}$ is the state space, $\mathcal{A}$ is the action space and
$\{P_t\}_{t=0}^T$ is a transition probability. $\{r_t\}_{t=0}^T : \mathcal{S} \times \mathcal{A} \;\to\; \mathbb{R}$ denotes the reward
received at time $t$.
A policy $\pi_t: \mathcal{S} \to \Delta(\mathcal{A})$ is a mapping from $s \in \mathcal{S}$ to a distribution over actions, usually parameterized by a neural network. The typical objective in reinforcement learning is to solve
\begin{equation}
\{\pi_t^*\}
\;=\;
\arg\max_{\{\pi_t\}}
\mathbb{E}_{\{\pi_t\}}
\Bigl[
  \sum_{t=0}^{T} r_t\bigl(s_t,a_t\bigr)
\Bigr]
\label{eq8}
\end{equation}
where $\mathbb{E}_{\{\pi_t\}}[\cdot]$ is the expectation induced by both
the policy $\pi$ and the transition probability. We take advantage of the inherent formulation of the diffusion model as an MDP to replace  $\pi_t$ with $p_t$ , since each policy corresponds directly to a denoising step.

\subsection{Entropy-Regularized MDPs}
We next present a reward-directed optimization objective in terms of a KL-regularized Markov decision process, tailored for diffusion models, reframing reverse diffusion as a policy under soft-reward constraints.
In entropy-regularized MDPs, we consider the following regularized objective:
\begin{equation}
\{\pi_t^*\}
\;=\;
\arg\max_{\{\pi_t\}}
\mathbb{E}_{\{\pi_t\}}
\Biggl[
  \sum_{t=0}^{T} r_t\bigl(s_t,a_t\bigr)
  \;-\;
  \alpha\,\mathrm{KL}\bigl(\pi_t(\cdot \mid s_t),\,\pi_t'(\cdot \mid s_t)\bigr)
\Biggr]
\label{eq10}
\end{equation}
where $\pi': \mathcal{S} \to \Delta(\mathcal{A})$ is a certain reference
policy. The $\arg\max$ solution is often called a set of soft optimal
policies. Compared to the standard objective in Eq.~\ref{eq8}, here we add KL terms against the reference policy, which later will be defined by the pretrained DDPM. This addition aims to ensure that soft optimal
policies align closely with the reference policies. Here, these reference policies correspond to the pre-trained diffusion models, as we aim to maintain
similarity between the fine-tuned and pre-trained models. 
Now, by reformulating the entropy-regularized objective into diffusion model notation, the objective function in Eq.~\ref{eq10} reduces to the following:
\begin{equation}
\{p_t^*\}_t
\;=\;
\arg\max_{\{\,p_t \}^{1}_{t=T+1}}
\Bigl[
  \mathbb{E}_{\{p_t\}}\bigl[r(x_0)\bigr]
  \;-\;
  \alpha
  \sum_{t=T+1}^{1}
  \mathbb{E}_{\{p_t\}}
  \Bigl[
    \mathrm{KL}\bigl(
      p_t(\cdot \mid x_t)
      \,\big\|\, 
      p_t^{\mathrm{pre}}(\cdot \mid x_t)
    \bigr)
  \Bigr]
\Bigr]
\label{eq11}
\end{equation}
where the expectation $\mathbb{E}_{\{p_t\}}[\cdot]$ is taken under the reverse process distribution $\prod_{t=T+1}^{1} p_t(x_{t-1} \mid x_t)$, and $\mathbb{E}_{\{p_t\}}\bigl[r(x_0)\bigr]$ is basically the future expected reward at $t = 0$. We set this argmax over the reverse process as an objective function in fine-tuning diffusion models. It defines the optimal reverse diffusion process $\{p_t^*\}$ that maximizes the expected reward over the generated samples while staying close to the pre-trained model via KL regularization. This balances two competing goals, namely exploiting high-reward regions by increasing likelihood of generating high-performance designs while staying close to pre-trained DDPM to avoid drifting into unrealistic or infeasible regions of design space.

\subsection{Soft Optimal Value Function}
To determine optimal reverse transitions within the entropy-regularized MDP framework, we consider the policy $p_t^*\bigl(x_{t-1}\mid x_t\bigr)$ that maximizes the expected cumulative reward while remaining close to the pre-trained reverse dynamics. This enables the characterization of the optimal policy in terms of the soft-value function, which serves as a smooth approximation of the maximum future reward and enables gradient-free guidance during both fine-tuning and inference.
Based on the results from Uehara et al.~\cite{uehara2024bridging}, who provided a theorem by induction showing that, from \( t = 0 \) to \( t = T+1 \), the form of the optimal policy \( p_t^*(x_{t-1} \mid x_t) \) is related to the soft-value function:
\begin{equation}
p_t^*\bigl(x_{t-1}\mid x_t\bigr)
:=\;
\frac{\exp\bigl(v_{t-1}(x_{t-1})/\alpha\bigr)\,
      p_t^{\mathrm{pre}}\bigl(x_{t-1}\mid x_t\bigr)}
     {\exp\bigl(v_t(x_t)/\alpha\bigr)}
\label{eq13}
\end{equation}
where \( v_t(\cdot) \) denotes the soft-value function. Here, \( p_t^*\bigl(x_{t-1} \mid x_t\bigr) \) assigns higher probability to transitions toward earlier states \( x_{t-1} \) with greater soft-value estimates, as given by \( v_{t-1}(x_{t-1}) \). The reference policy \( p_t^{\mathrm{pre}} \) serves as a baseline to ensure that the learned transitions remain close to the pre-trained distribution, while the numerator introduces a directional bias toward samples that improve the reward. The scaling factor \( \alpha \) controls the trade-off between greedy behavior and exploration.
To effectively reweight the pre-trained diffusion model in the direction of higher reward, the marginal distribution at time \( t \), defined as $ p_t^\star(x_t)
= \int \Bigl[\prod_{k=T+1}^{t+1} p_k^\star(x_k \!\mid\! x_{k-1})\Bigr] \, \mathrm{d}x_{t+1:T}
$ , can be recursively expressed as follows:

\begin{equation}
p_t^\star(x_t)
= \frac{\exp\bigl(v_t(x_t)/\alpha\bigr)\,p_t^{\mathrm{pre}}(x_t)}{C}
\label{eq12}
\end{equation}
where \(C\) is the normalization constant. The soft optimal value function can be correlated with the pre-trained diffusion model by recursively building from the soft Bellman equation \cite{uehara2024bridgingmodelbasedoptimizationgenerative}:
\begin{equation}
\exp\Bigl(\tfrac{v_t(x_t)}{\alpha}\Bigr)
\;=\;
\int
  \exp\Bigl(\tfrac{v_{t-1}(x_{t-1})}{\alpha}\Bigr)\,
  p_t^{\mathrm{pre}}(x_{t-1}\mid x_t)\,dx_{t-1}
\;=\;\dots
\;=\;
\mathbb{E}_{\{p_t^{\mathrm{pre}}\}}
\Bigl[
  \exp\!\Bigl(\tfrac{r(x_0)}{\alpha}\Bigr)
  \;\Big|\;
  x_t
\Bigr]
\label{eq16}
\end{equation}
where \(\mathbb{E}_{\{p_t^{\mathrm{pre}}\}}[\cdot \mid x_t]\) means 
\(\mathbb{E}_{x_0 \sim p_1^{\mathrm{pre}}(x_1),\,\dots,\,x_{t-1} \sim p_{t-1}^{\mathrm{pre}}(x_t)}[\cdot \mid x_t].\) 
Unlike standard value functions that compute expected reward directly, the soft formulation introduces the exponential mapping to promote high-reward outcomes more aggressively, while still maintaining smoothness and differentiability with respect to the probability distribution. This gradient-free mechanism is useful for guiding both fine-tuning and inference toward high-reward regions in the design space.

\subsection{Reward-Weighted MLE}\label{mlesec}
To solve Equation~\ref{eq11} and optimize reward within a supervised‐learning framework, Norouzi et al.\ \cite{norouzi2016reward} proposed a reward‐based maximum‐likelihood approach. Reward‐weighted maximum‐likelihood estimation (MLE) is widely used in reinforcement learning \cite{peng2019advantageweightedregressionsimplescalable,peng2019advantage,fan2023dpok}, and variants have been applied to diffusion models \cite{zhang2023controllablediffusionmodelsrewardguided,uehara2024understanding}. We adopt reward‐weighted MLE because it closely matches the structure of reward‐based importance sampling and helps preserve the pre‐trained distribution to keep the design within the feasible design domain. Below, we write the optimal policy \( p_t^* \) in the context of diffusion model, obtained by maximizing the value-weighted log-likelihood:

\begin{equation}
p_t^*(\cdot)
\;=\;
\arg\max_{p_t \in \Pi_t}
\;\mathbb{E}_{\,x_{t-1}\sim p_{t-1}^{\mathrm{pre}}(\cdot\mid x_t),\,x_t\sim u_t}
\Bigl[
  \exp\bigl(\tfrac{v_{t-1}(x_{t-1})}{\alpha}\bigr)\,
  \log\,p_t\bigl(x_{t-1}\mid x_t\bigr)
\Bigr]
\label{eq14}
\end{equation}
where \(u_t\) is a roll-in policy defined by the current policy at time \(t\), modeled as a Gaussian distribution with mean parameterized by a neural network with parameters \(\theta\). 
The value function at each timestep is not known, but we can use Equation~\ref{eq16} at timestep \(x_{t-1}\) and substitute it into Equation~\ref{eq14}. We can then rewrite the optimal policy with the reward at \(t = 0\) as follows:
\begin{equation}
p_t^*(\cdot)
\;=\;
\arg\max_{p_t \in \Pi_t}
\;\mathbb{E}_{%
  x_0 \sim p_1^{\mathrm{pre}}(x_1),\;
  \dots,\;
  x_{t-1} \sim p_{t-1}^{\mathrm{pre}}(x_t),\;
  x_t \sim u_t
}
\biggl[
  \exp \!\bigl(\tfrac{r(x_0)}{\alpha}\bigr)\,
  \log p_t(x_{t-1}\mid x_t)
\biggr]
\label{eq:rewarded-log-prob}
\end{equation}
where the negative log likelihood, $ - \log p_t(x_{t-1}\mid x_t)$, is similar to the original loss function. The updated loss function of the reward-weighted MLE is simply a weighted version of the original pre-trained diffusion model( Equation \ref{eq:ddpm_elbo} and Equation \ref{lddpm} ) as follows:
\begin{equation}
L(\theta) =
\sum_{t = T+1}^{1} \,\sum_{i=1}^{m}
\Biggl[
  w^{(i,t)}\|
     \epsilon_{t}^{(i,t)} 
     \;-\;
     \frac{
       \epsilon\bigl(x_{t}^{(i,t)},\,T - t;\,\theta_{\mathrm{pre}}\bigr)
       \;-\;
       \epsilon\bigl(x_{t}^{(i,t)},\,T - t;\,\theta\bigr)
     }{
       \{\sigma_t^\diamond\}^2
     }
  \Bigr\|_{2}^{2}
\Biggr]\Biggl.
\label{eq20}
\end{equation}
where $\epsilon\bigl(x_{t}^{(i,t)},\,T - t;\,\theta\bigr)$ is the learnable parameter during fine-tuning and $
w^{(i,t)} \;=\; \exp\!\Bigl(\frac{r(x_0^{(i,t)})}{\alpha}\Bigr)$ is the weight parameter that affects the loss function. Algorithm~\ref{alg:reward_weighted_mle} illustrates the reward-weighted fine-tuning procedure, where $m$ denotes the batch size, $\gamma$ represents the learning rate, and $\alpha \in \mathbb{R}^{+}$ is a scaling parameter. 

\begin{figure}[htbp]

    \centering
   
    \tikzstyle{block} = [rectangle, rounded corners, minimum width=1cm, minimum height=1.5cm, text centered, draw=black, fill=blue!20]
    \tikzstyle{block2} = [rectangle, rounded corners, minimum width=0.8cm, minimum height=1.3cm, text centered, draw=black, fill=white!20]
    \tikzstyle{blockgray} = [rectangle, rounded corners, minimum width=1cm, minimum height=1.5cm, text centered, draw=black, fill=black!10]
    \tikzstyle{tildeblock} = [rectangle, rounded corners, minimum width=1cm, minimum height=1.5cm, text centered, draw=black, fill=blue!20]
    \tikzstyle{arrow} = [thick,->,>=stealth]
    \tikzstyle{dashedarrow} = [thick, dashed,->,>=stealth]
    \tikzstyle{curvedarrow} = [thick,->,>=stealth, bend right=30]
    \tikzstyle{curvedarrowrev} = [thick,dashed,->,>=stealth, bend left=30]
    \tikzstyle{dottedarrow} = [thick, dotted,., fill=black]

 \resizebox{1.1\textwidth}{!}{
    \begin{tikzpicture}[node distance=2.5cm]
        \node (xT) [block2] {$x_T$};
        \node (xTm1) [block2, right=2.5cm of xT] {$x_{T-1}$};
        \node (xTm2) [block2, right=2.7cm of xTm1] {$x_{T-2}$};
        \node (xtildeT2) [block2, below=1.5cm of xTm2] {$\tilde{x}_{T-2}$};
        \node (xtilde1) [block2, right=3.7cm of xtildeT2] {$\tilde{x}_{1}$};
        \node (x1) [block2, above=1.5cm of xtilde1] {$x_1$};
        \node (xtilde0) [block2, right=2.6cm of xtilde1] {$\tilde{x}_{0}$};
        \node (x0) [blockgray, right=2.5cm of x1] {$x_0$};
        \node at (8.55,0.1) {$\cdots$}; 
        \node at (17,0) {\text{RDD}}; 

        \draw [dashedarrow] ($(xTm1.west)+(0,+0.2)$) -- node[above] {$q(x_T | x_{T-1})$} ($(xT.east)+(0,+0.2)$);
        \draw [dashedarrow] ($(x1.west)+(0,+0.2)$) -- node[above] {$ q(x_{2}|x_{1})$} ($(xTm2.east)+(1,+0.2)$);
        \draw [dashedarrow] ($(xTm2.west)+(0,+0.2)$) -- node[above] {$q(x_{T-1} | x_{T-2})$} ($(xTm1.east)+(0,+0.2)$);
        \draw [dashedarrow] ($(x0.west)+(0,+0.2)$) -- node[above] {$q(x_1 | x_0)$} ($(x1.east)+(0,+0.2)$);

        \draw [arrow] (xT) -- node[below] {$P_{\theta}(x_{T-1} | x_T)$} (xTm1);
        \draw [arrow] ($(xTm1)+(0.7,-0.5)$) -- node[below] {$ $} ($(xtildeT2)+(-0.65,0)$);
        \draw [arrow] (xtildeT2) -- node[left] {$\displaystyle w^{(i,t)}$} (xTm2);
        \draw [arrow] (xTm1) -- node[below] {$P_{\theta}$} (xTm2);
        \draw[arrow]  ($(xTm2.east)+(1,0)$)-- node[below] {$P_{\theta} $} (x1);
        \draw [arrow] (xtilde1) -- node[left] {$\displaystyle w^{(i,t)}$} (x1);
        \draw [arrow] (9.2,-.4) -- node[right] {$ $} ($(xtilde1)+(-0.4,0)$);
        \draw [arrow] ($(x1)+(+0.5,-0.5)$) -- node[below] {$ $} ($(xtilde0)+(-0.4,0)$);
        \draw [arrow] (xtilde0) -- node[left] {$\displaystyle w^{(i,t)}$} (x0);
        \draw [arrow] (x1) -- node[below] {$P_{\theta}$} (x0);
    \end{tikzpicture}
}

 \rule{0pt}{0.5cm}
\resizebox{1.125\textwidth}{!}{
    \begin{tikzpicture}[node distance=2.8cm]
        \node (xT) [block2,below=5cm of xT] {$x_T$};
        \node (xTm1) [block2, right=2.8cm of xT] {$x_{T-1}$};
        \node (xTm2) [block2, right=2.9cm of xTm1] {$x_{T-2}$};
        \node (x1) [block2, right=3.6cm of xTm2] {$x_1$};
        \node (x0) [blockgray, right=2.8cm of x1] {$x_0$};
        \node at  ($(xTm2.east) + (0.5,-0.1)$){$\cdots$}; 
         \node at (18,-6.6) {\text{Vanilla}}; 

        \draw [dashedarrow]($(xTm1.west) + (0,+0.2)$) -- ($(xT.east) + (0,+0.2)$);
        \draw [dashedarrow]($(xTm2.west) + (0,+0.2)$) -- node[above] {$ q(x_{T-1}|x_{T-2}) $}($(xTm1.east) + (0,+0.2)$);
        \draw [dashedarrow] ($(x1.west) + (0,+0.2)$)  -- node[above] {$q(x_{2}|x_{1})$} ($(xTm2.east) + (1,+0.17)$);
        \draw [dashedarrow] ($(x0.west)+(0,+0.2)$) -- node[above] {$ q(x_{1}|x_{0}) $} ($(x1.east)+(0,+0.2)$);

        \draw [arrow] (xT) -- node[below] {$ $} (xTm1);
        \draw [arrow] (xTm1) -- node[below] {$ p_{\theta}(x_{T-2} | x_{T-1})$} (xTm2);
        \draw[arrow] ($(xTm2.east) + (1,0)$) -- node[below] {$p_{\theta}(x_{1}|x_{2})$} (x1);
        \draw [arrow] (x1) -- node[below] {$p_{\theta}(x_{0} | x_{1})$} (x0);
    \end{tikzpicture}
    }

    \caption{Graphical model of the vanilla DDPM (bottom) and reward-directed diffusion (RDD, top). In RDD,  $\tilde{x}_{T-t}$ is the intermediate pretrained step and ${x}_{T-t}$ is the model directed towards high reward values.}
    \label{schem}
\end{figure}
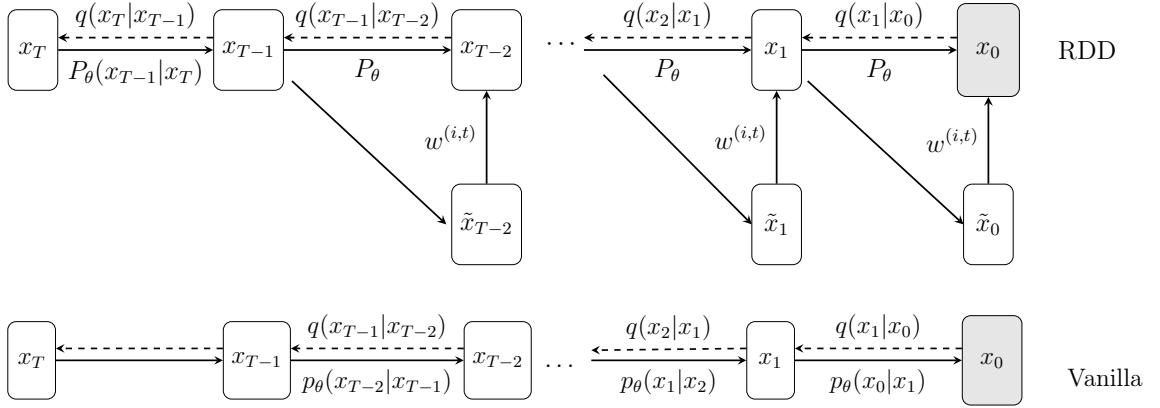

Fig.~\ref{schem} illustrates the forward and reverse processes in reward-directed diffusion (RDD) fine-tuning and highlights the key differences from the standard DDPM. In this figure, \( \tilde{x}_{T-t} \) denotes the intermediate state before reward guidance at step \( T - t \) in the reverse process, while \( x_{T-t} \) represents the corresponding state after being guided toward higher reward values. Reward values correspond to the performance of the design: for airfoil design, the reward is defined as the normalized lift-to-drag ratio ($C_l/C_d$), while for ship design, it is given by the scaled negative resistance to encourage drag reduction. Note that the reward value is reduced by a feasibility penalty in both design cases. In other words, \( r(x_0) = \hat{r}(x_0) - \hat{g}(x_0) \), where \( \hat{g} \) represents the feasibility penalty. \( \hat{g} \) is not a model but rather a constraint function over the design variables. For the airfoil design case, it penalizes designs with coordinates beyond the \([0, 1]\) range or those that induce self-intersections. For the ship design case, feasibility constraints are defined for all 44 parameters, following the definitions in \cite{bagazinski2023shipgen}.

\begin{algorithm}[ht]
\caption{Fine-tuning pre-trained model using reward-weighted MLE}
\label{alg:reward_weighted_mle}
\begin{algorithmic}[1]
\STATE Initialize: $\theta_1 = \theta_{\mathrm{pre}}$
\FOR{$s \in \{1, \dots, S\}$}
  \FOR{$k \in \{T+1, \dots, 1\}$}
    \STATE \textbf{Collect $m$ samples} 
      $\{\,(x_k^{(i,t)})_{i=1,k=T}^{m,0}\,\}$
      from a policy 
      $p_{T+1}(\cdot \mid \cdot;\,\theta_s),\dots,p_{t+1}(\cdot \mid \cdot;\,\theta_s),
       p_t^{\mathrm{pre}}(\cdot \mid \cdot),\dots,p_1^{\mathrm{pre}}(\cdot \mid \cdot).$
  \ENDFOR
  \STATE Update $\theta_s$ to $\theta_{s+1}$ as follows:
  \[
    \theta_{s+1}
=\;
\theta_s
\;-\;
\gamma\,
\nabla_\theta
\sum_{t,i}
w^{(i,t)}\,
\frac{\|x_{t-1}^{(i,t)} - \mu(x_t^{(i,t)},t;\theta)\|^2}
     {(\sigma_t)^2}
\Big\vert_{\theta=\theta_s}
    \tag{24}
  \]
\ENDFOR
\STATE \textbf{Output:} $\{\,p_t(\cdot \mid \cdot;\,\theta_s)\}_{t=T+1}^{1}$
\end{algorithmic}
\end{algorithm}

\subsection{Reward-based Importance Sampling}\label{rbis}
Reward-based or Soft Value-based Decoding in Diffusion model (SVDD) closely matches the reward-weighted fin-tuning in terms of the objective function \cite{li2024derivative}. Both are designed to generate designs with maximum rewards from an optimal policy using soft-value function. The difference is that reward-based sampling performs only sampling from the optimal policy, as defined in Equation \ref{eq11}. Unlike fine-tuning, which collects samples from intermediate states using a roll-in policy and re-trains with a weighted loss based on the expected reward at $t = 0$, the reward-based sampling process does not require additional training. Since our distribution closely aligns with the optimal policy through fine-tuning, we use a soft value function approximation for inference instead of collecting samples from intermediate states and retraining the posterior distribution based on reward feedback at the final state. A soft value function $v_{t}(x_t) $ at each step of the reverse process is correlated with the expected reward at $ t = 0  $ by Equation \ref{eq16}. Instead of using the expected reward at $t = 0$, we approximate the soft value function by the expected reward at $x_0$. In other words, $\widehat{v}_{t}(x_t) = r(\mathbb{E}[\,x_0 \mid x_t\,]) $. We rely on the posterior mean approximation using Equation \ref{eq2} to obtain the estimate $\mathbb{E}[\,x_0 \mid x_t\,]$ of $x_0$ at time step $t$:

\begin{equation}
\mathbb{E}[x_0 \mid x_t]
  = \frac{x_t - \sqrt{1 - \bar{\alpha}_t}\,\epsilon}
         {\sqrt{\bar{\alpha}_t}}.
\end{equation}
Replacement of $\epsilon$ with the prediction from the fine-tuned model ($\epsilon_\theta(x_t, t)$) yields the following:
\begin{equation}
\widehat{x}_0
  = \frac{x_t - \sqrt{1 - \bar{\alpha}_t}\,\epsilon_{\theta}(x_t, t)}
         {\sqrt{\bar{\alpha}_t}}.
\label{exptd_x0}
\end{equation}
Using Eq.~\ref{exptd_x0}, we approximate the expected final design $x_0$ in each intermediate state   $x_t$  by the posterior mean estimator, which depends on the predicted noise $\epsilon_{\theta}(x_t, t)$ from the fine-tuned diffusion model. 
This approximation enables directional sampling without backward passes or reward gradients and forms the basis for reward-based importance sampling.  At each reverse diffusion step, we sample multiple candidate states, score them using the soft value approximation, and select one using weighted sampling. 

Algorithm~\ref{alg:svdd} shows the process for a single trajectory of reward-based importance sampling \cite{li2024derivative}. At each reverse-diffusion step, we draw a set of candidate samples and assign weights using the soft-value approximation evaluated at $\widehat{x}_0$. These weights define a categorical distribution (denoted $\mathrm{Categorical}$ in Algorithm~\ref{alg:svdd}), which assigns higher probability to samples with higher rewards.
In the current implementation, we perform 1000 parallel trajectories. This means that for each run, we generate 1000 high-reward design candidates to choose from. The number of trajectories was selected based on memory capacity constraints. It directly impacts the number of candidates ($M$) that can be sampled, evaluated, and selected at each step using one-draw sampling. For the A100 GPU with 40 GB of memory used in this study, we limit $M$ to values less than 11, which we find to be a suitable trade-off between the number of trajectories and the number of samples per step, considering memory constraints. The motivation for developing an iterative reward-based framework was the limitation of the value of $M$, which restricts the sampling of high-reward designs. By fine-tuning based on the same principle of the soft optimal policy, we distribute the cost of alignment between the training and inference phases to achieve rewards beyond the data distribution.
After describing the optimization framework, we now turn our attention to its practical application. To evaluate the proposed reward-directed diffusion model, we consider two canonical design problems, namely 2D airfoil and 3D ship hull optimization, parameterized in a manner to enable generative modeling and reward evaluation.

\begin{algorithm}[tb]
\caption{Reward-directed importance sampling}
\label{alg:svdd}
\begin{algorithmic}[1]
  \REQUIRE Fine-tuned DDPM $\{p_t^{\mathrm{RDD}}\}_{t=T}^0$, posterior mean approximation for soft value function$\{\hat{v}_t\}_{t=0}^T$, and hyperparameter $\alpha\in\mathbb{R}$
  \FOR{$t = T,\,T-1,\,\dots,1$}
    \STATE Sample proposals
    \[
      \{x_{t-1}^{\langle m\rangle}\}_{m=1}^M \;\sim\; p_{t-1}^{\mathrm{RDD}}(\,\cdot\mid x_t)\,.
    \]
    \FOR{$m=1,\dots,M$}
      \STATE Score each candidate with soft value
      \[
        w_{t-1}^{\langle m\rangle}
        \;\gets\;
        \exp\!\bigl(\hat v_{t-1}(x_{t-1}^{\langle m\rangle})/\alpha\bigr).
      \]
    \ENDFOR
    \STATE Draw index $\zeta_{t-1}$ using one-draw sampling
    \[
      \zeta_{t-1} \;\sim\;
      \mathrm{Categorical}\!\Bigl(\bigl\{
        w_{t-1}^{\langle m\rangle} \big/ \sum_{j=1}^M w_{t-1}^{\langle j\rangle}
      \bigr\}_{m=1}^M\Bigr),
    \]
    and set
    \[
      x_{t-1} \;\gets\; x_{t-1}^{\langle \zeta_{t-1}\rangle}.
    \]
  \ENDFOR
  \STATE \textbf{Output:} $x_0$.
\end{algorithmic}
\end{algorithm}

\section{Data Processing and Geometry Representation}
\begin{figure}[h!]
    \centering
    \includegraphics[width=0.7\textwidth]{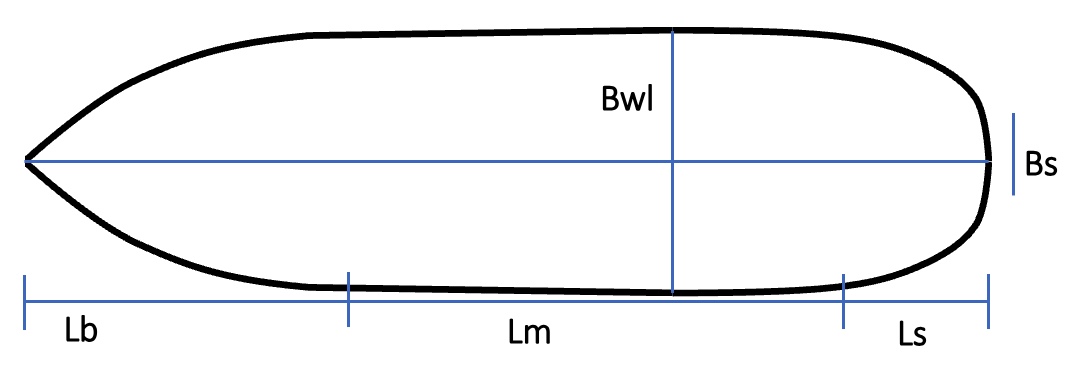} 
    \caption{Graphical illustration of selected parameters used in the general hull form parametrization} 
    \label{cross} 
\end{figure}

We utilize a parametric geometry representation, which consists of a collection of design parameters organized in a tabular or vectorized format. Each row corresponds to a single design, while each column represents a specific design parameter. We consider two cases: airfoil design in 2D space and ship hull design in 3D space. For the airfoil case, each row contains 384 parameters, representing 192 points with $(x, y)$ coordinates that define the 2D shape, along with their corresponding \(C_l/C_d\) value. 
We use 38,000 airfoil designs from an augmented version of the UIUC airfoil dataset presented by Nobari et al.~\cite{heyrani2021pcdgan}. 
The data are generated under incompressible flow conditions with a Reynolds number of \(1.8 \times 10^6\), Mach number \(Ma = 0.01\), and angle of attack \(\alpha = 0^\circ\). For the ship hull design task, we use a tabular dataset that is a combination of data available in Ship-D dataset \cite{bagazinski2023ship}, C-ShipGen \cite{bagazinski2024c}, and data generated during the pre-training phase of the DDPM. We select a subset of 10,000 data points from this combined dataset for reward-based training. We briefly describe the method used for parametric hull generation and its practical implementation. The approach defines the geometry of the hull through a set of non-dimensional input parameters that are then used to construct continuous mathematical representations of the hull. This method provides clear boundaries for the geometry of the hull and provides adequate design freedom for design space exploration. Each row of the data contains 44 design variables that construct the full hull geometry. Fig.~\ref{cross} shows some of the parameters used to define the general hull form, which is parameterized by the following: LOA (Length Overall); \(L_b\) and \(L_s\) (bow and stern taper lengths); \(B_d\) and \(D_d\) (beam and depth at the deck); \(B_s\) (beam at the stern); and WL (waterline depth).
In addition to the general hull form, the geometry representation includes midbody cross-section parameters, bow form parameters, bulbous bow parameters, and stern form parameters. The geometric parameters are expressed in dimensionless form, scaled by the ship’s length overall (LOA) and depth (D$_{d}$) when applicable. The parameters are represented as a vector $\mathbf{p} = [p_1, p_2, \dots, p_n]$ defined as:
\begin{equation}
L_b = p_1 \cdot LOA, \quad L_s = p_2 \cdot LOA, \quad B_d = p_3 \cdot LOA, \quad D_d = p_4 \cdot LOA,
\label{eq28}
\end{equation}
and 
\begin{equation}
B_s = p_5 \cdot \frac{B_d}{2}, \quad WL = p_6 \cdot D_d
\label{eq29}
\end{equation}

\begin{figure}[h!]
    \centering
    \includegraphics[width=0.85\textwidth]{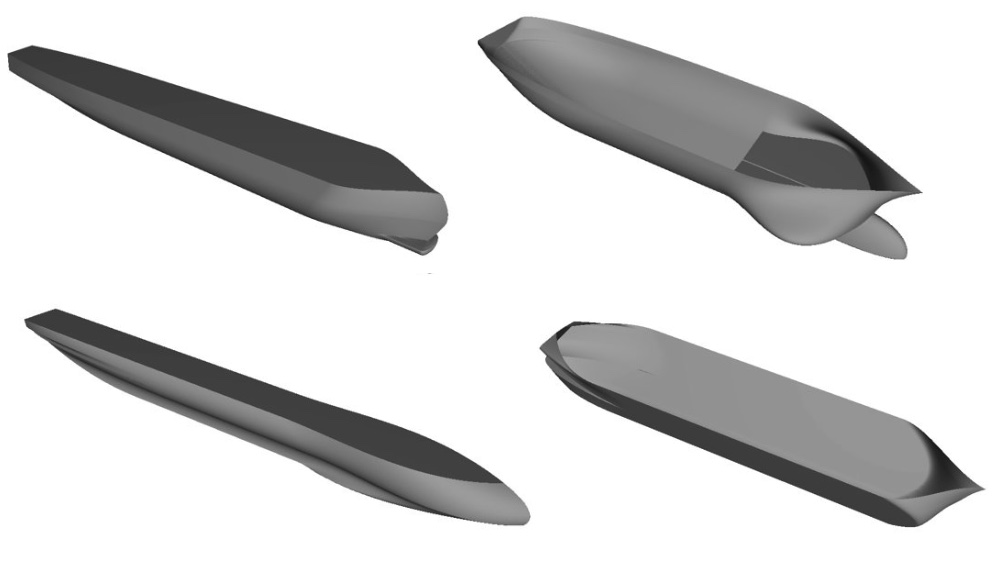} 
    \caption{Representative 3D ship hull designs from the dataset.} 
    \label{stlexample} 
\end{figure}

These scaling relations convert the input parameters into dimensionally consistent parameters. Once all the functions are defined, a surface point cloud is generated. A triangulation algorithm is used to create a polygonal mesh, which can be exported in STL format for visual assessment. Figure~\ref{stlexample} shows the meshes in the STL file derived from the parameterization method. The reader is referred to Bagazinski et al.~\cite{bagazinski2023ship} for additional details.

Similar to the \(C_l/C_d\) value used for airfoil design, our aim is to optimize the performance of ship hulls generated using the diffusion model. We focus on minimizing calm-water resistance, and since each segment of the hull is defined by a precise mathematical representation, the hydrostatic and hydrodynamic properties can be directly computed through numerical integration. The wave-making resistance \( R_w \) can be estimated by an integral according to the Michell's integral \cite{tuck1989wave,michell1898xi}
\begin{equation}
R_w = \frac{A \rho g^2}{\pi U^2} \int_1^{\infty} (I^2 + J^2) \frac{\lambda^2}{\sqrt{\lambda^2 - 1}} \, d\lambda
\label{eq31}
\end{equation}
where \( A \), \( I \) and \( J \) are parameters relating to the surface normal across the hull and the direction of wave propagation, \( \rho \) is the water density, \( g \) is the gravitational acceleration, \( U \) is the velocity, \( \lambda \) is a dimensionless wavenumber parameter related to the ship’s geometry and wave pattern. To make the wave resistance \( R_w \) comparable across different ships, the wave resistance coefficient \( C_w \), defined as:

\begin{equation}
C_w = \frac{R_w}{\frac{1}{2} \rho U^2 \cdot \text{LOA}^2}
\label{eq32}
\end{equation}
where  \( \frac{1}{2} \rho U^2 \cdot \text{LOA}^2 \) is a reference dynamic pressure force.  
Arising from viscous effects between the ship's hull and the water, the frictional resistance is computed using the friction coefficient \(C_f\), which depends on the Reynolds number, is given as
\begin{equation}
C_f \;=\; \frac{0.075}{\bigl(\log(\mathrm{Re}) - 2\bigr)^2}
\label{eq33}
\end{equation}
and the frictional force can be expressed as
\begin{equation}
R_f \;=\; \tfrac12 \;C_f \;\rho\,U^2 \; S_{A,t} \;\text{LOA}^2
\label{eq34}
\end{equation}
where $S_{A,t}$ is a non-dimensionalized wetted surface area for the hull for each draft. The total resistance \( R_T \) is the estimated sum of wave resistance (\( R_w \)) and skin friction resistance (\( R_f \)). The aggregated \( R_T \) on eight different velocities (Froude numbers ranging from 0.1 to 0.45) and four different drafts (0.25, 0.33, 0.5, and 0.67) is considered the performance criterion and is used in this study as an objective function in the RDD model. An XGBoost surrogate model \cite{Chen_2016} is trained to accelerate resistance prediction, achieving an \( R^2 \) value of 0.985, which facilitates faster design generation and reward evaluation. While simplified physics are used for resistance calculations in this study, the framework can be extended to incorporate high-fidelity simulation data. Furthermore, the current framework optimizes resistance alone, without accounting for other critical aspects of ship design, such as maneuverability, stability, and seakeeping performance under varying sea states.
After having geometry representations and surrogate reward models, we proceed to empirical evaluation. This section presents results that compare the pre-trained DDPM with the reward-directed version, evaluating performance in terms of objective improvement and distributional shift.

\section{Results and Discussion}
To evaluate the proposed reward-directed diffusion framework, we consider two canonical shape optimization problems: the 2D airfoil and the 3D ship hull design. In both cases, a pre-trained denoising diffusion probabilistic model is fine-tuned using reward-weighted maximum likelihood estimation, followed by inference through directional sampling based on soft value functions. Specifically, we present the results after reward-based fine-tuning to generate optimized designs and compare them with those produced by sampling from a vanilla DDPM.

\subsection{2D Airfoil Generative Design}
In this section, we first present the result for the precision of the regression model that we trained as a reward model. An XGBoost is trained as a regression model to predict \(C_l/C_d\) value for each airfoil design. The coefficient of determination ($ R^2$ value) of the trained model is equal to 0.994 which shows high prediction performance. This regression model was used to act as a reward model during fine-tuning. Decision tree architectures are well suited to structured engineering design data, as they are easy to train and often outperform neural networks on tabular inputs \cite{shwartz2022tabular}. Their simplicity in achieving a strong baseline, combined with a superior handling of skewed or heavy-tailed feature distributions and other dataset irregularities~\cite{mcelfresh2023neural}, makes tree-based models a reliable choice for engineering predictions.

\begin{figure}[H]
    \centering
    \includegraphics[width= .82\textwidth]{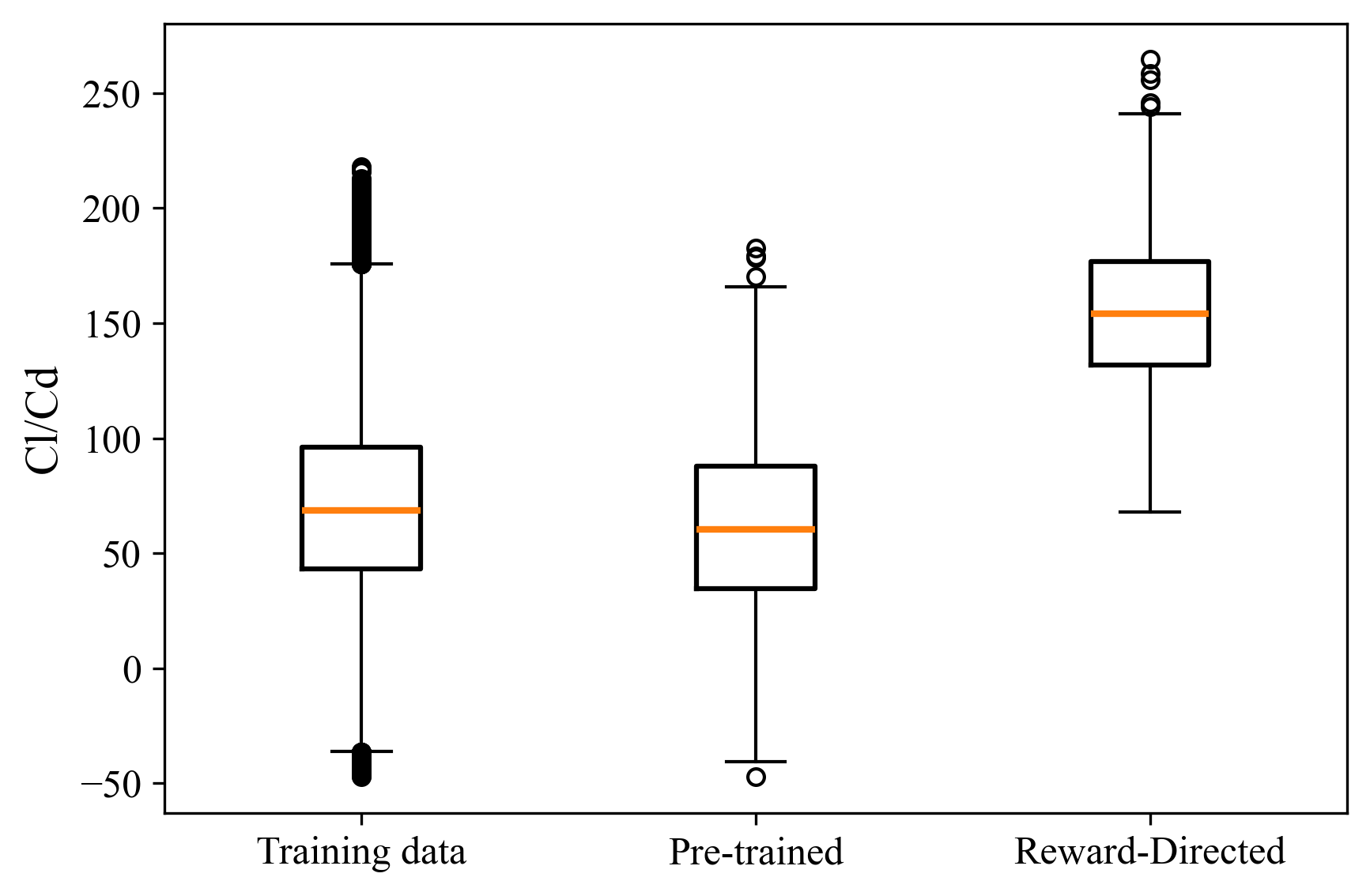} 
    \caption{Boxplot comparing $Cl/Cd$ distributions from training data, pre-trained DDPM, and reward-directed diffusion in the 2D airfoil design task. Reward-directed diffusion shifts the distribution toward higher-performance regions.} 
    \label{Boxplot1} 
\end{figure}

Fig.~\ref{Boxplot1} shows the design distribution in the form of a box plot for the training data, the pre-trained DDPM, and the reward-directed diffusion. 
Each box represents the interquartile range (IQR), with the median marked by a horizontal line, and whiskers extending to 1.5 $\times$ IQR. Outliers beyond this range are plotted as individual points. 
The DDPM is pre-trained to generate designs that are stochastically similar to the airfoils, with no optimization applied. Therefore, no improvement in terms of \(C_l/C_d\) is observed in the generated designs. However, the reward-directed diffusion model generates airfoil designs with a higher median \(C_l/C_d\) value and more consistent aerodynamic performance than those found in the initial dataset. Note that the reward-directed data shown in Fig.~\ref{Boxplot1} are based on the complete framework, which includes both fine-tuning and sampling from the fine-tuned model.

\begin{figure}[H]
    \centering
    \includegraphics[width= .82\textwidth]{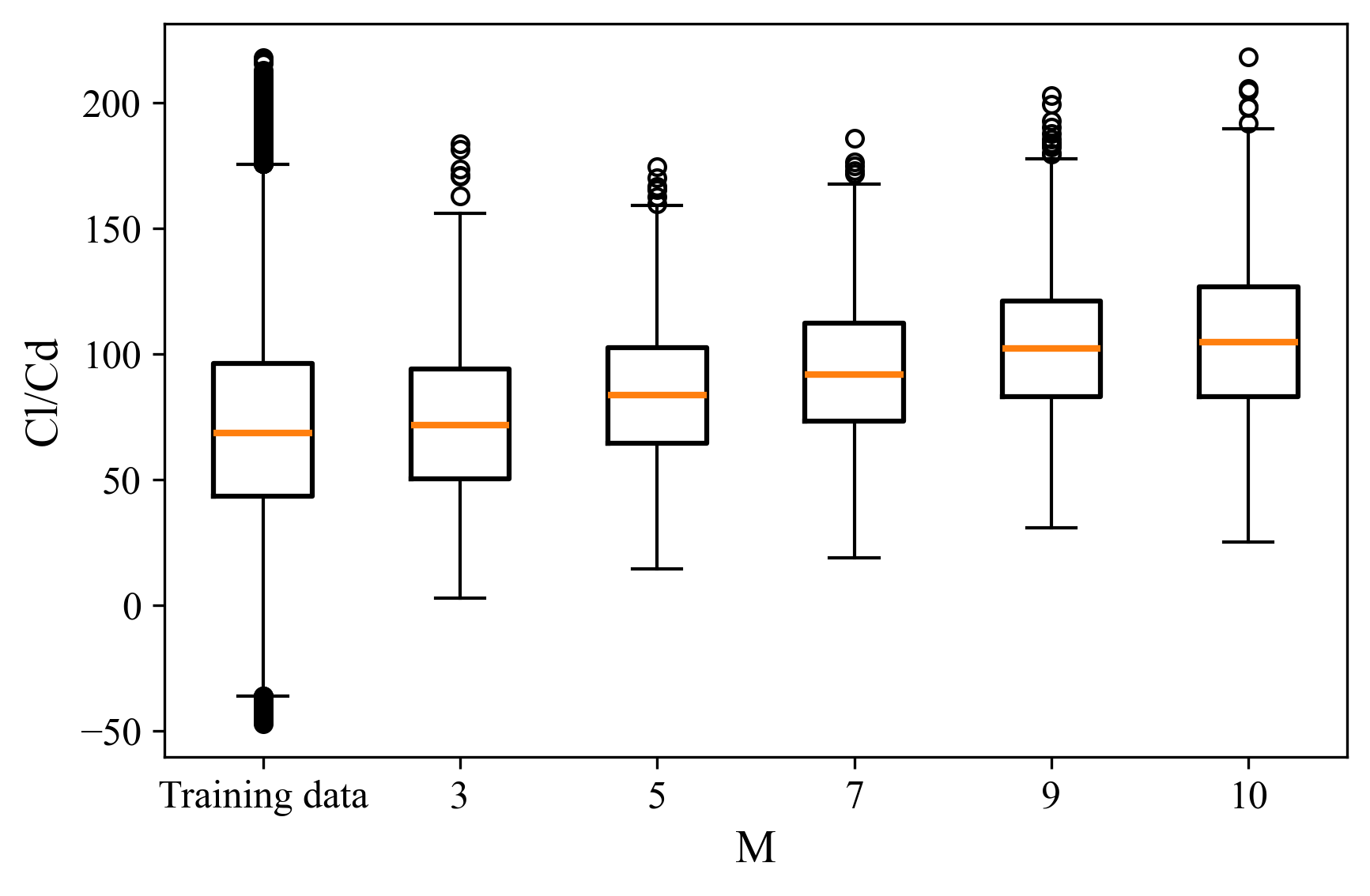} 
    \caption{Distribution of sampled designs from pre-trained DDPM using reward-based importance sampling, for different values of sampling budget \(M\). As \(M\)increases, higher-performing samples are more frequently selected.} 
    \label{Boxplotairfoil2} 
\end{figure}

Fig.~\ref{Boxplotairfoil2}, on the other hand, shows the training data and the results of the reward-based sampling of the pre-trained model, rather than the fine-tuned model. This experiment is conducted to further study the impact of the iterative reward-directed approach. Here, \(M\) denotes the number of candidates sampled and evaluated using the soft value function at each step of the reverse process. Increasing \(M\) to 10 and possibly beyond leads to higher reward values. However, higher values of \(M\) generally increase memory cost during inference, which can limit scalability depending on the available hardware resources. Even with higher values of \(M\), the generated samples do not exceed the reward range observed in the training data. While the mean reward of the generated samples increases, they remain within the bounds of the training data distribution.

\begin{figure}[h!]
    \centering
    \includegraphics[width= .82\textwidth]{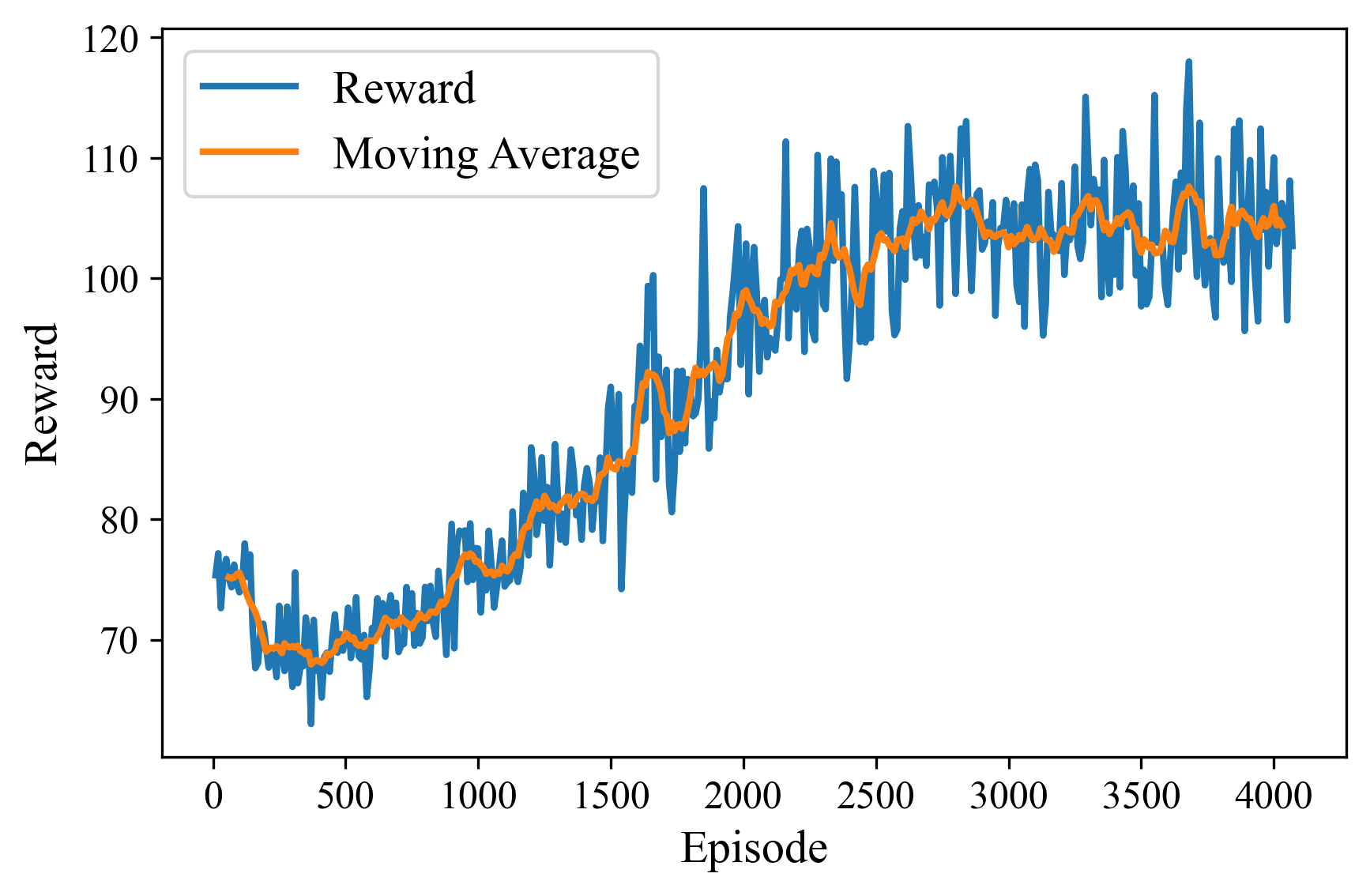} 
    \caption{Reward history during fine-tuning 2D model using reward-weighted MLE. A consistent increase in expected reward over training episodes is observed. } 
    \label{RLrewardAirfoil} 
\end{figure}

We further fine-tune the model using the method described in Section~\ref{mlesec} and Algorithm~\ref{alg:reward_weighted_mle} to guide the model parameters towards higher-performance designs. The unnormalized reward history during fine-tuning is shown in Fig.~\ref{RLrewardAirfoil}, demonstrating how the rewards at \(x_0\) shift toward higher expected values. Fine-tuning progressively increases the reward of the generated samples, with values exceeding 110 in later episodes.

\begin{figure}[H]
    \centering
    \includegraphics[width= .82\textwidth]{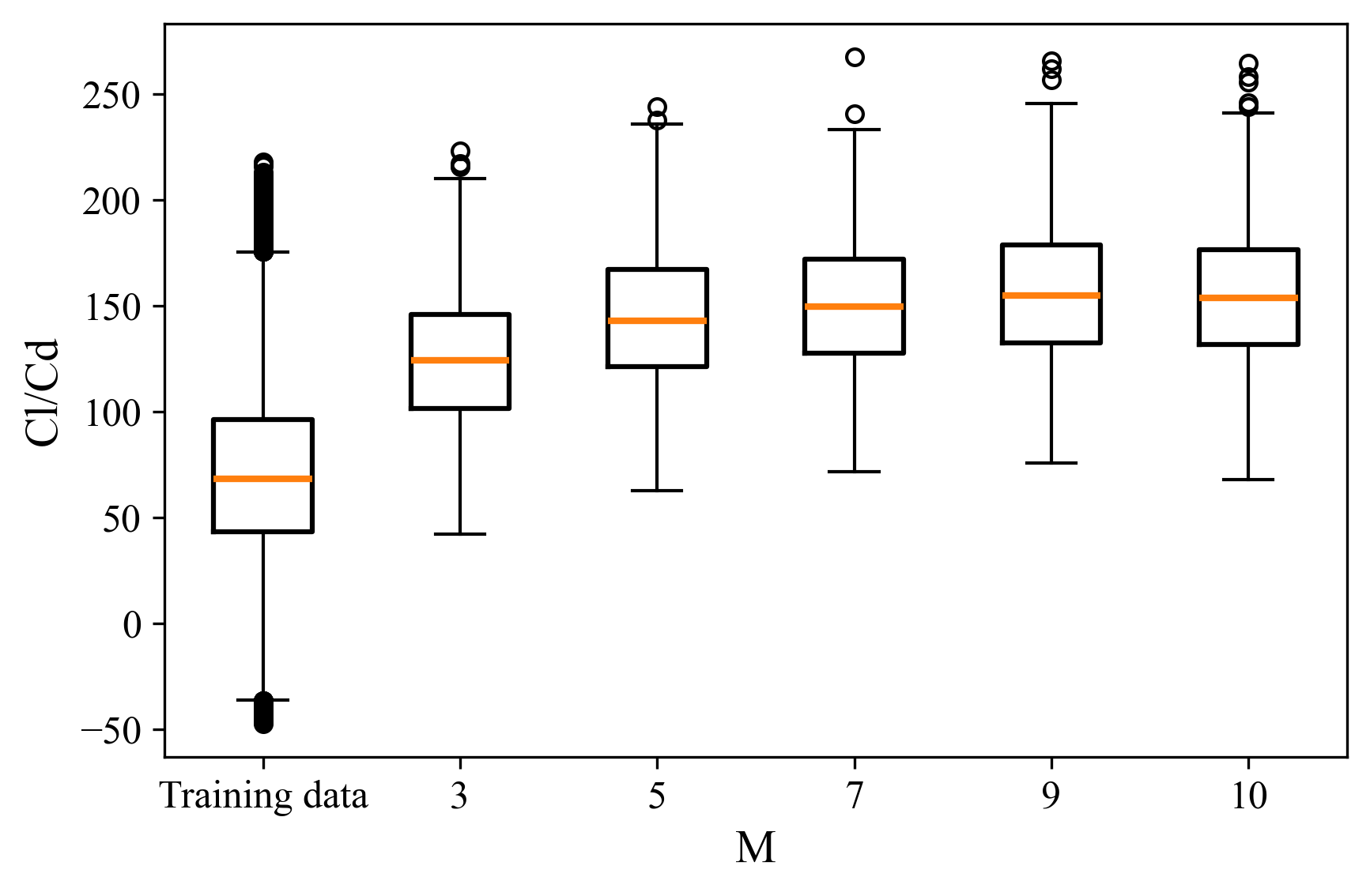} 
    \caption{Distribution of the training data and sampled designs from the reward-directed diffusion model across different values of \(M\)} 
    \label{Boxplotairfoil3} 
\end{figure}

Fig.~\ref{Boxplotairfoil3} shows the box plot of the sampled data from the reward-directed diffusion model. Note that, as mentioned in the methodology section, the process iteratively follows the steps described in Sections~\ref{mlesec} and~\ref{rbis}. What is interesting is that the generated samples now exceed the performance of the training data, which is significant given that the airfoil training data already reside in an optimized design space. This makes it difficult to find values for \(C_l/C_d\) beyond those in the original dataset.

\begin{figure}[H]
    \centering
    \includegraphics[width= .82\textwidth]{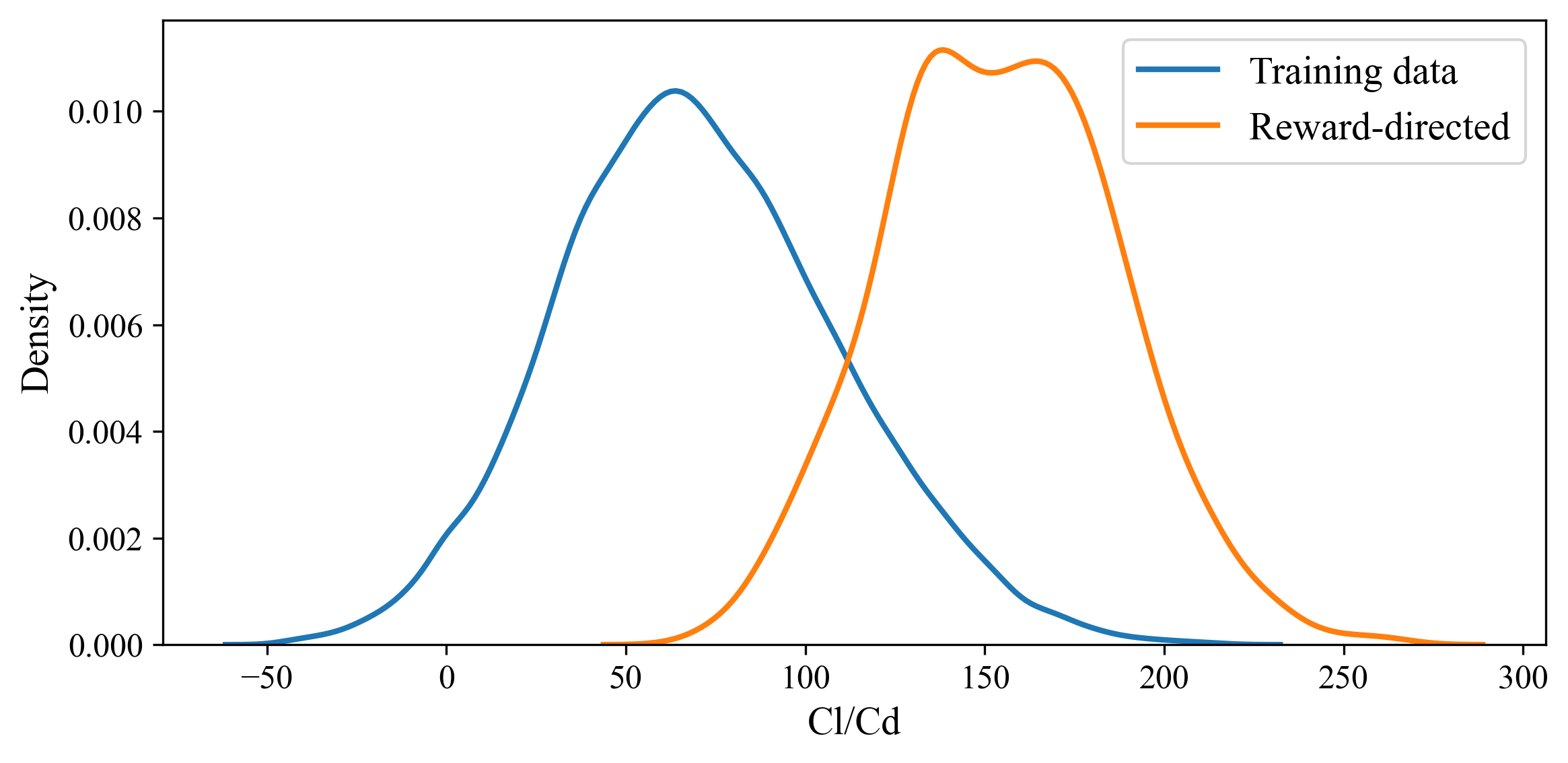} 
    \caption{Probability density function of \(C_l/C_d\) values comparing the training data and sample airfoil designs generated by the reward-directed diffusion model, showing a positive shift toward higher-performance designs} 
    \label{airfoilkde} 
\end{figure}

\begin{figure}[H]
    \centering
    \includegraphics[width= .76\textwidth]{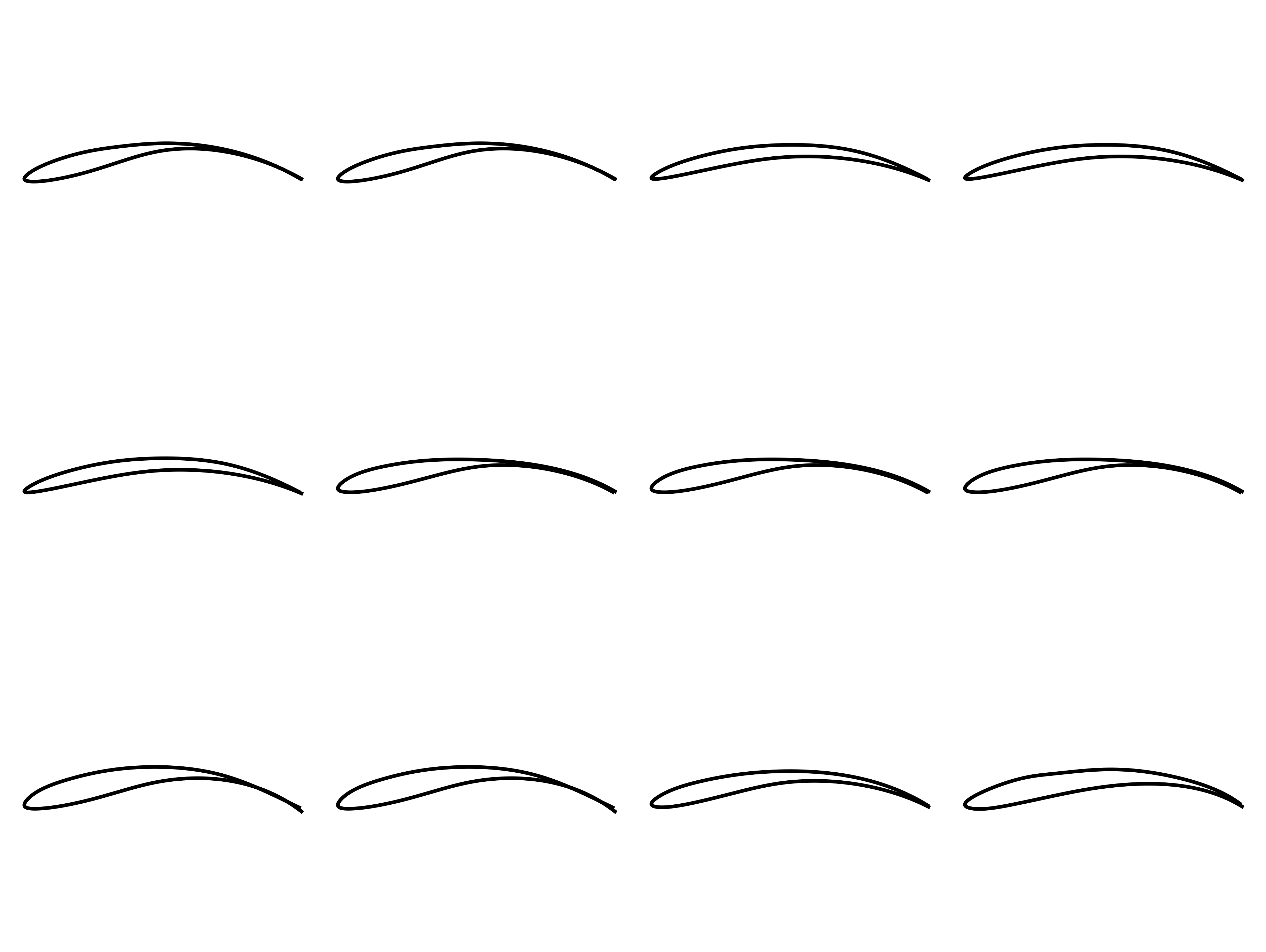} 
    \caption{High-performance airfoil designs generated via reward-directed diffusion. Cubic spline interpolation is applied to design coordinates. } 
    \label{Airfoil} 
\end{figure}

The probability density function of the data is shown in Fig.~\ref{airfoilkde} to illustrate the distribution of the training and sampled data, as well as the positive shift in \(C_l/C_d\) values, both in the mean and in the higher-performance region of the distribution. 
We further use reward-directed diffusion to generate a collection of high-reward samples, which are shown in Fig.~\ref{Airfoil}. We consider a cubic spline to connect the points and create a smooth curve. 


\begin{figure}[h!]
    \centering
    \includegraphics[width=.89\textwidth]{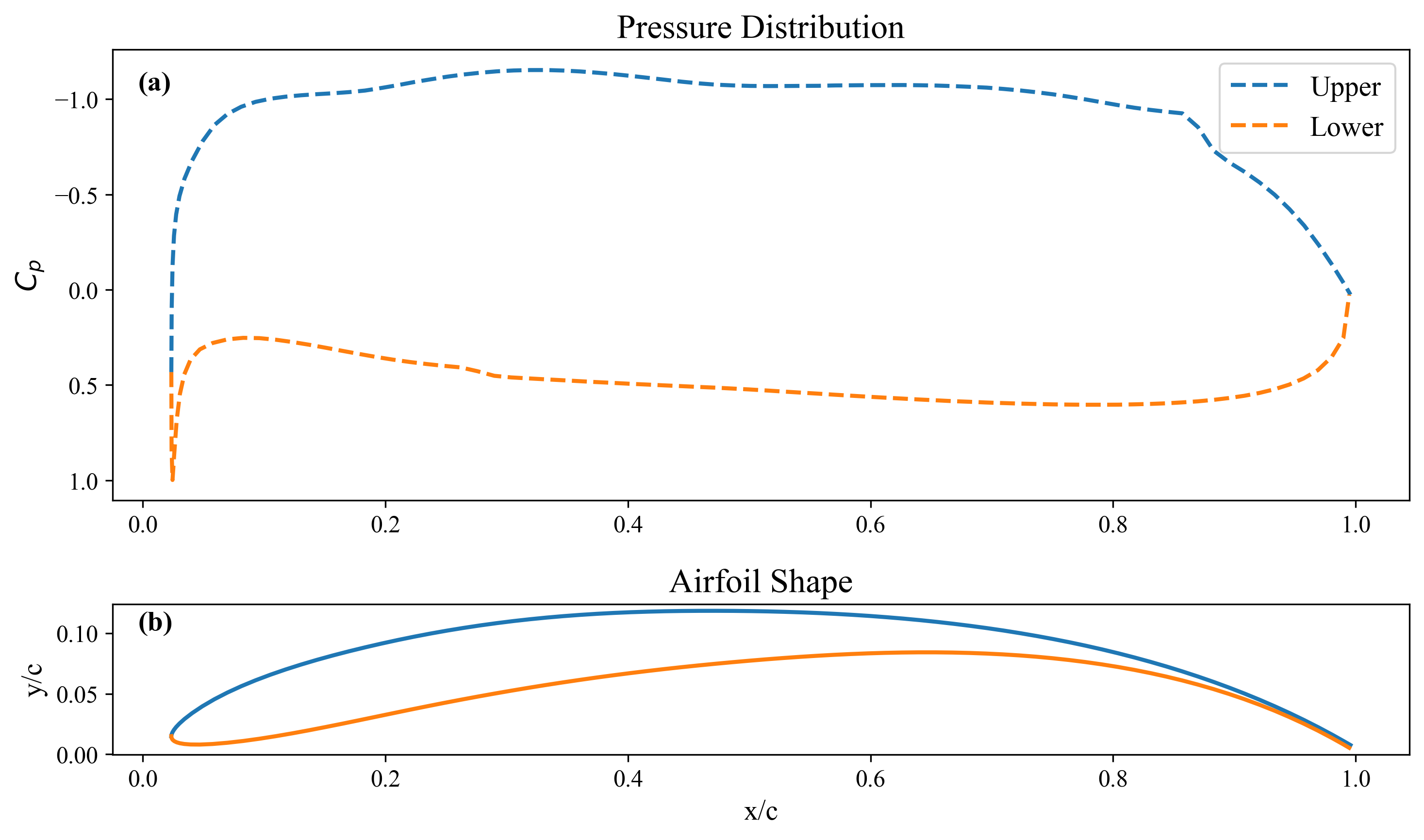} 
    \caption{Airfoil shape and pressure distribution along the airfoil surface for one of the well-performing designs (\(c \) is the chord length). This airfoil has a \(C_l/C_d\) value of 275.56.} 
    \label{Airfoil_CP} 
\end{figure}

These samples are validated using XFOIL simulations\footnote{\url{https://web.mit.edu/drela/Public/web/xfoil/}}. A detailed profile of one of these well-performing designs along with its corresponding pressure distribution is shown in Fig. \ref{Airfoil_CP}. This airfoil has a \(C_l/C_d\) value of 275.56 which is well beyond the highest value in the training data. 
We next apply the same framework to the more complex problem of optimizing the 3D hull of a ship. This not only evaluates the scalability of the method, but also demonstrates its effectiveness in higher-dimensional and hydrodynamically richer design spaces.

\subsection{3D Ship Hull Design Optimization}
This section presents the results for the optimization of the 3D hull of a ship. As in the previous section, we aim to find high-reward (i.e., reduced drag) designs while preserving naturalness, meaning that the generated designs remain in proximity to the initial dataset in order to retain important properties such as hydrostatic stability. We describe the sampling process that guides the resistance to decrease in order to increase the reward. Since the reward is defined as the linear scaled negative negative of resistance, minimizing the resistance corresponds to maximizing the reward.

\begin{figure}[h!]
    \centering
    \includegraphics[width= .82\textwidth]{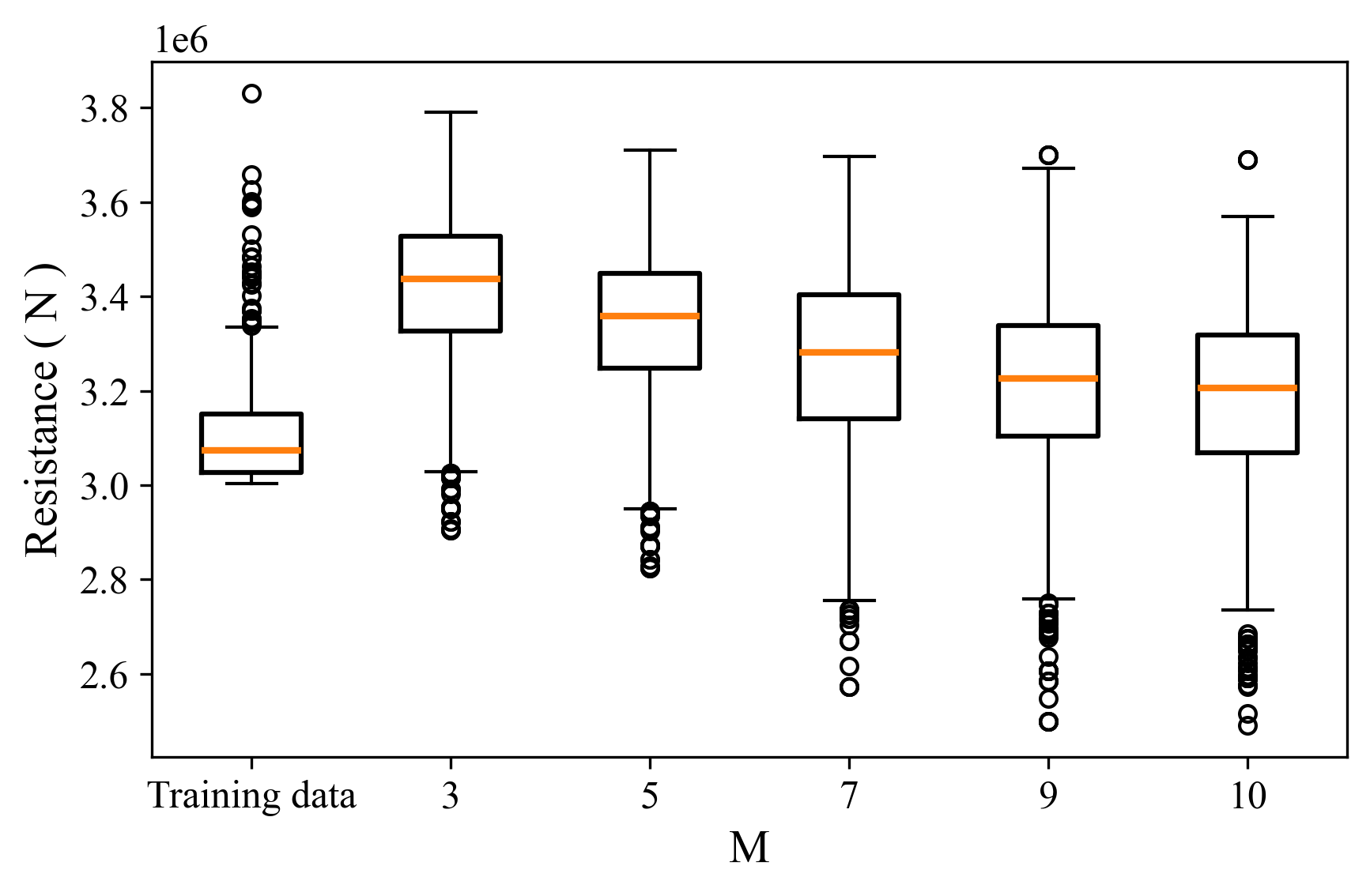} 
    \caption{Box plot of calm-water resistance values for training data and samples generated from the pre-trained model using reward-based importance sampling in the 3D ship hull case} 
    \label{Shipsampling} 
\end{figure}

\begin{figure}[h!]
    \centering
    \includegraphics[width= .84\textwidth]{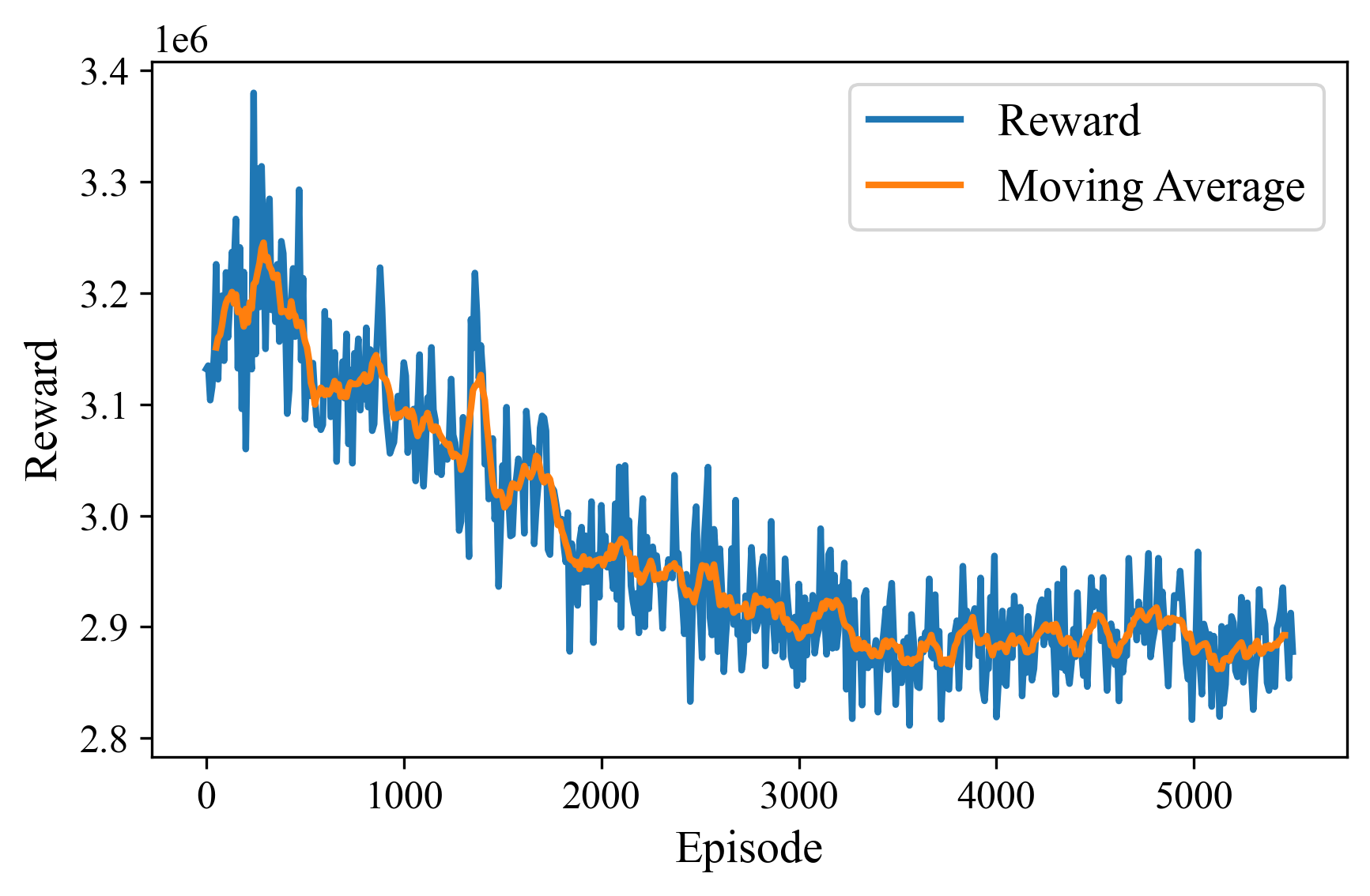} 
    \caption{Reward history during fine-tuning for 3D Ship hull design using reward-weighted MLE } 
    \label{RLrewardship} 
\end{figure}

Fig.~\ref{Shipsampling} shows the box plot of calm-water resistance across generated hull designs. The training data and the optimized designs generated using reward-based importance sampling from the pre-trained model are presented in the figure. 
The boxes represent the central 50\% of the data (IQR), with medians shown as horizontal lines and whiskers extending up to 1.5$\times$ IQR.  The reward-guided model produces hulls with significantly lower median resistance and reduced variability, demonstrating both performance gains and distributional robustness compared to the baseline diffusion model.
Increasing the number of design candidates (\(M\)) at each step of the reverse process leads to a lower resistance in the generated samples. We use \(M = 10\) to remain consistent with the 2D test case. Unlike the 2D case, reward-based importance sampling from the pre-trained model in the 3D case yields optimized designs beyond the training data, which is particularly seen at higher values of \(M\).
As in the 2D case, we perform fine-tuning using the methods described in Section~\ref{mlesec} to address the memory capacity limitations that arise when increasing the number of candidates \(M\). The reward, as mentioned earlier, is a linearly scaled negative cost to enable minimization rather than maximization, but is converted to resistance values for illustrative purposes in Fig.~\ref{RLrewardship}.

\begin{figure}[h!]
    \centering
    \includegraphics[width= .84\textwidth]{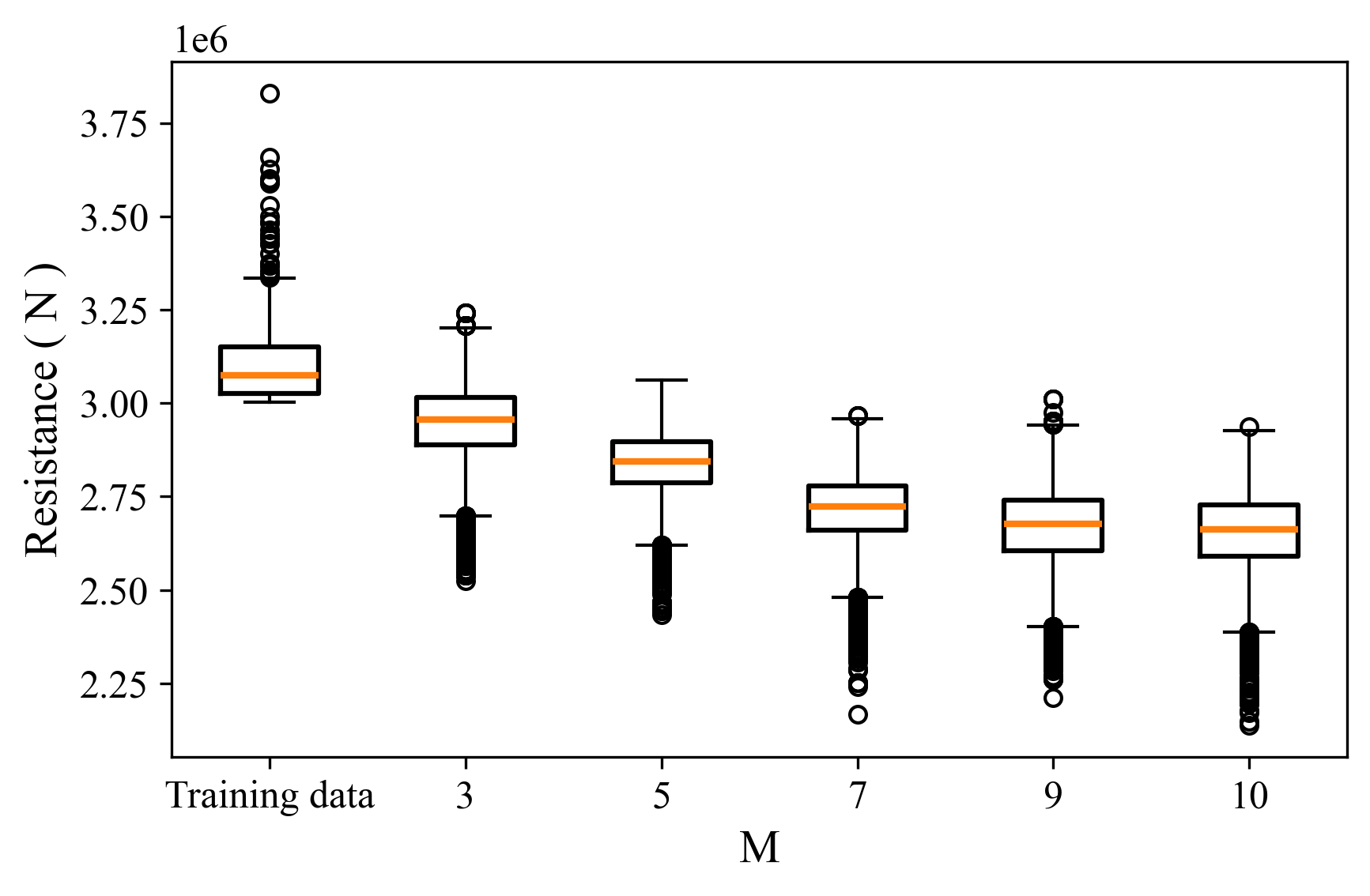} 
    \caption{Box plot of resistance values for samples generated from the reward-directed diffusion model in the 3D ship hull case. The model produces significantly lower resistance designs.} 
    \label{shiptuned} 
\end{figure}

Fig.~\ref{shiptuned} shows the samples generated after fine-tuning using reward-weighted MLE, followed by reward-based importance sampling using the method described in Section~\ref{rbis} which we call the reward-directed diffusion model. Although reward-based sampling from the pre-trained model in the 3D ship hull case shown in Fig.~\ref{Shipsampling} already produces optimized designs beyond the training data, further fine-tuning and iterative soft value guidance reduce the resistance of the generated samples even more, achieving over 25\% reduction in resistance. This iterative approach leverages training-phase improvements in the soft value to reduce memory requirements while further directing the samples toward the optimized design space.

\begin{figure}[h!]
    \centering
    \includegraphics[width= .85\textwidth]{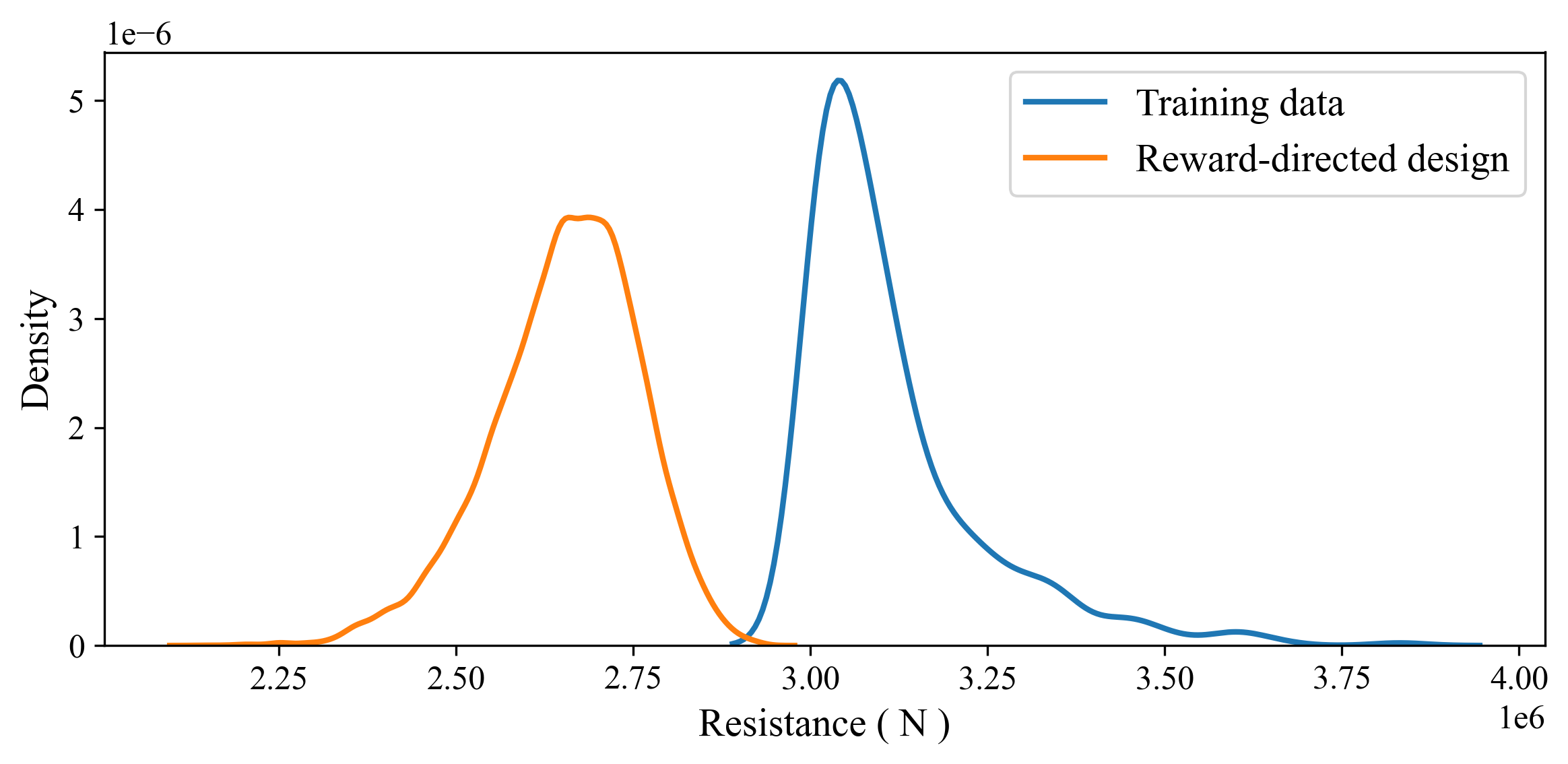} 
    \caption{Probability density of resistance for training and sampled 3D hull designs from reward-directed diffusion, showing a larger shift toward reduced drag than in the 2D case} 
    \label{shipkde} 
\end{figure}

\begin{figure}[h!]
    \centering
    \includegraphics[width= \textwidth]{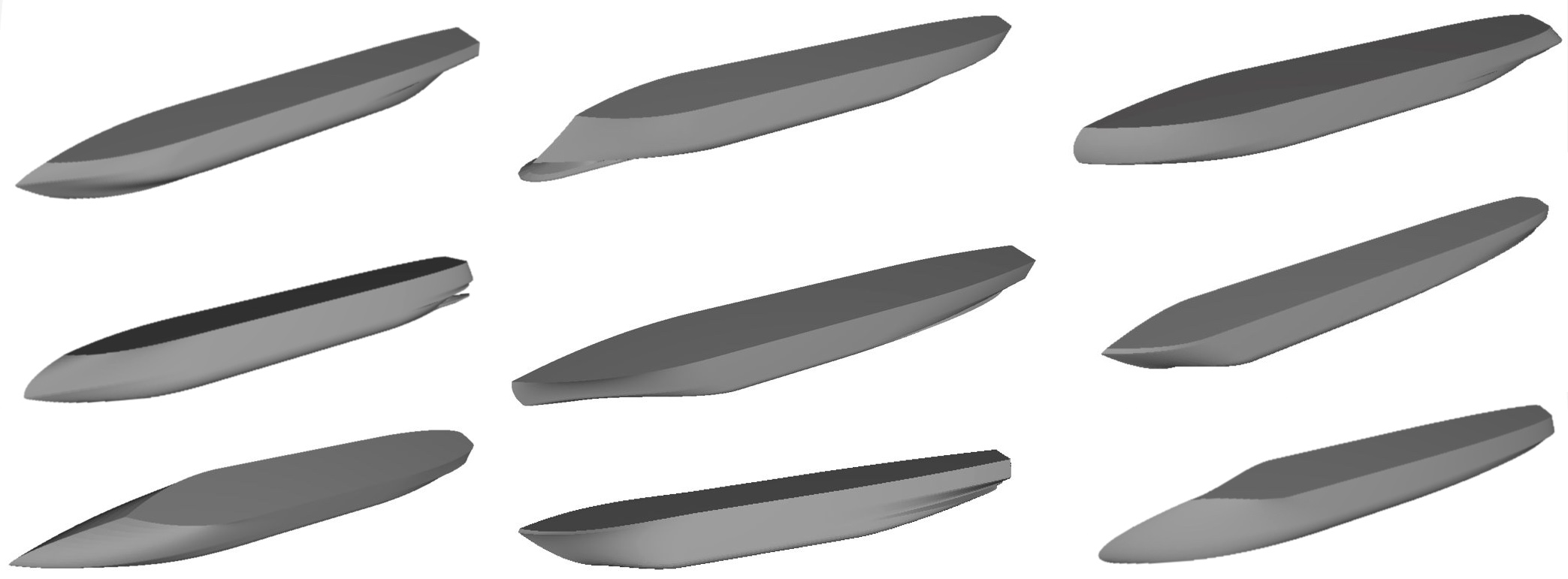} 
    \caption{Example ship hull geometries generated by reward-directed diffusion. Visual inspection shows smooth and streamlined shapes.} 
    \label{shipsamplestl} 
\end{figure}

The probability density function of the resistance data is shown in Fig.~\ref{shipkde} to illustrate the distribution of the training and generated sample designs. A more pronounced shift toward reduced drag values is observed, both in the mean and in the lower tail of the distribution, compared to the airfoil case, as the ship hull training data did not reside in an optimized design space. A collection of low-resistance ship hull designs is shown in Fig.~\ref{shipsamplestl}.

\begin{figure}[h!]
    \centering
    \begin{subfigure}[b]{0.48\textwidth}
        \centering
        \includegraphics[width=\textwidth]{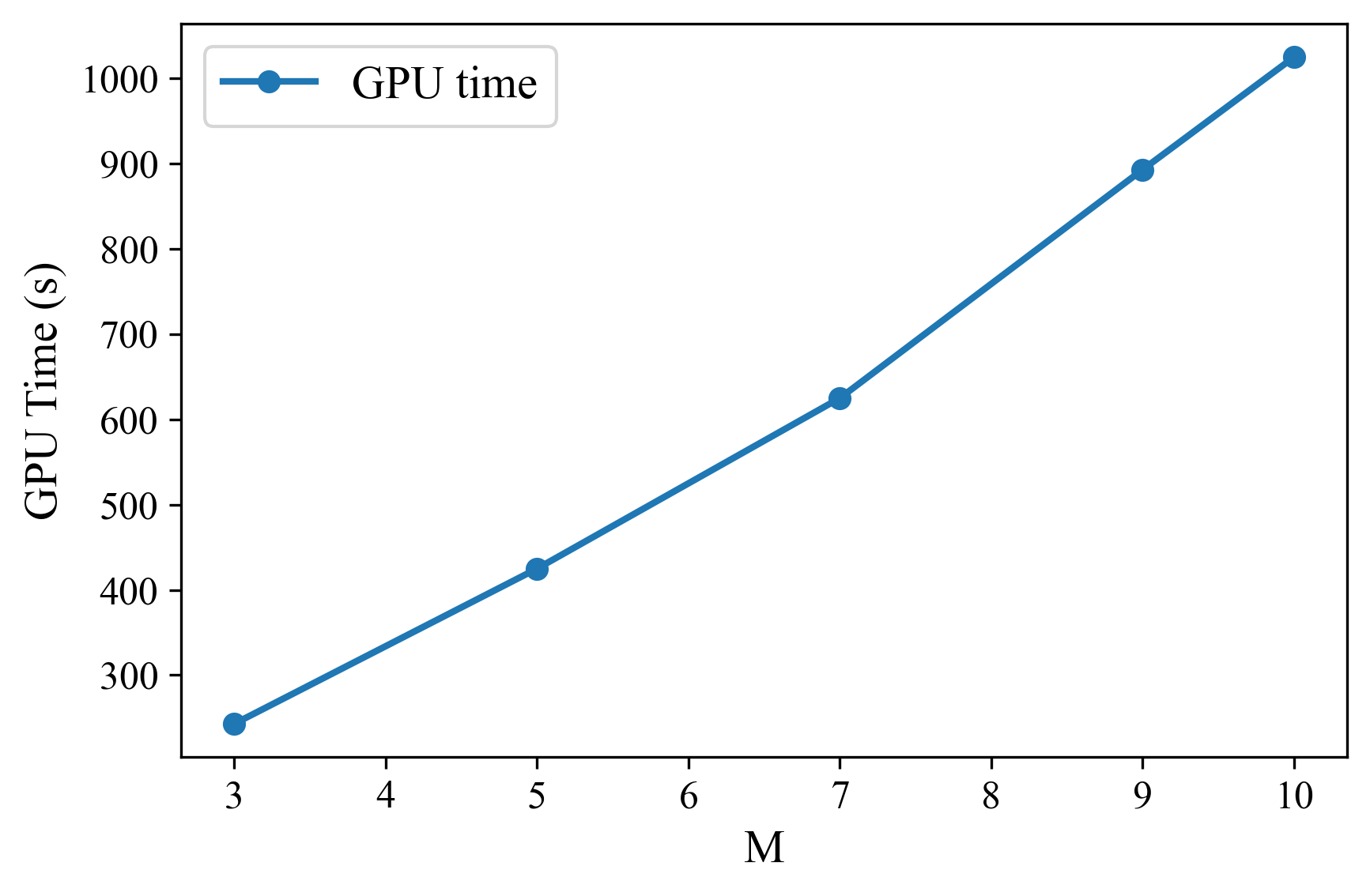}
        \caption{2D Airfoil design}
        \label{fig:gpu_ship}
    \end{subfigure}
    \hfill
    \begin{subfigure}[b]{0.48\textwidth}
        \centering
        \includegraphics[width=\textwidth]{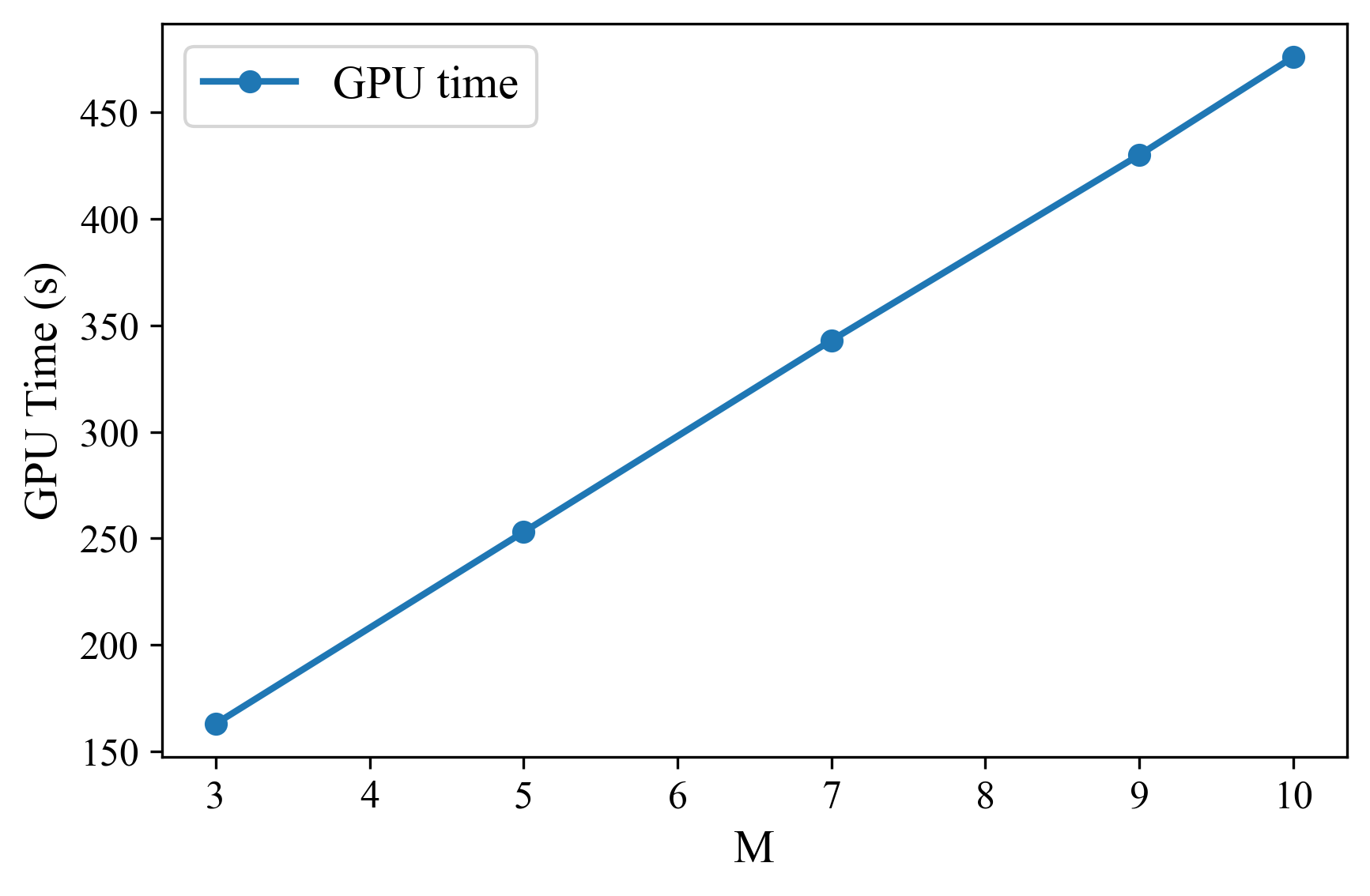}
        \caption{3D ship hull design}
        \label{fig:gpu_airfoil}
    \end{subfigure}
    \caption{GPU computation time versus number of candidates \(M\) for (a) 2D airfoil and (b) 3D ship hull designs, measured on an A100 GPU. The results show an approximately linear increase in computation time as \(M\) increases. }
    \label{fig:gpu_times_combined}
\end{figure}

By fine‑tuning the diffusion model with a reward‑weighted MLE and then applying reward‑based importance sampling, we align the model towards high‑performance designs while respecting memory constraints (e.g., the limit on \(M\)). This approach mitigates the memory bottleneck of direct sampling and enables the model to approach the optimal policy without expensive gradient computations.
Fig.~\ref{fig:gpu_times_combined} shows the GPU computation time as a function of the number of candidates \(M\) for both the 2D airfoil and 3D ship hull design cases. As \(M\) increases from 3 to 10, the processing time increases nearly linearly, reflecting the proportional cost of evaluating more samples per reverse diffusion step using reward-based importance sampling. Our framework reduces the dependency on increasing the number of candidates \(M\) and the inference time to generate optimized designs.

\subsubsection{Validation of 3D Ship Hull Designs}
The commercial software MAXSURF \cite{maxsurf2013} is used to verify the hydrostatic, stability, and powering details of the hulls generated from the reward-directed diffusion model. 
The generated hull from the reward-directed diffusion model was exported as an STL file. Inside the MAXSURF modeler, the mesh was inspected for watertightness, automatically segmented into chine strips, and the NURBS surface of best fit was generated by the Mesh-to-Surface routine of the software. This surface then served as the single parametric model that was passed on to the hydrostatic, large-angle stability, and resistance modules. All analyzes were performed with the density set to \(\SI{1}{\text{t}/\text{m}^{-3}}\)and with the trim locked to the even keel condition.

\begin{figure}[h!]
  \centering
  \begin{subfigure}[b]{0.45\textwidth}
    \centering
    \includegraphics[width=\textwidth]{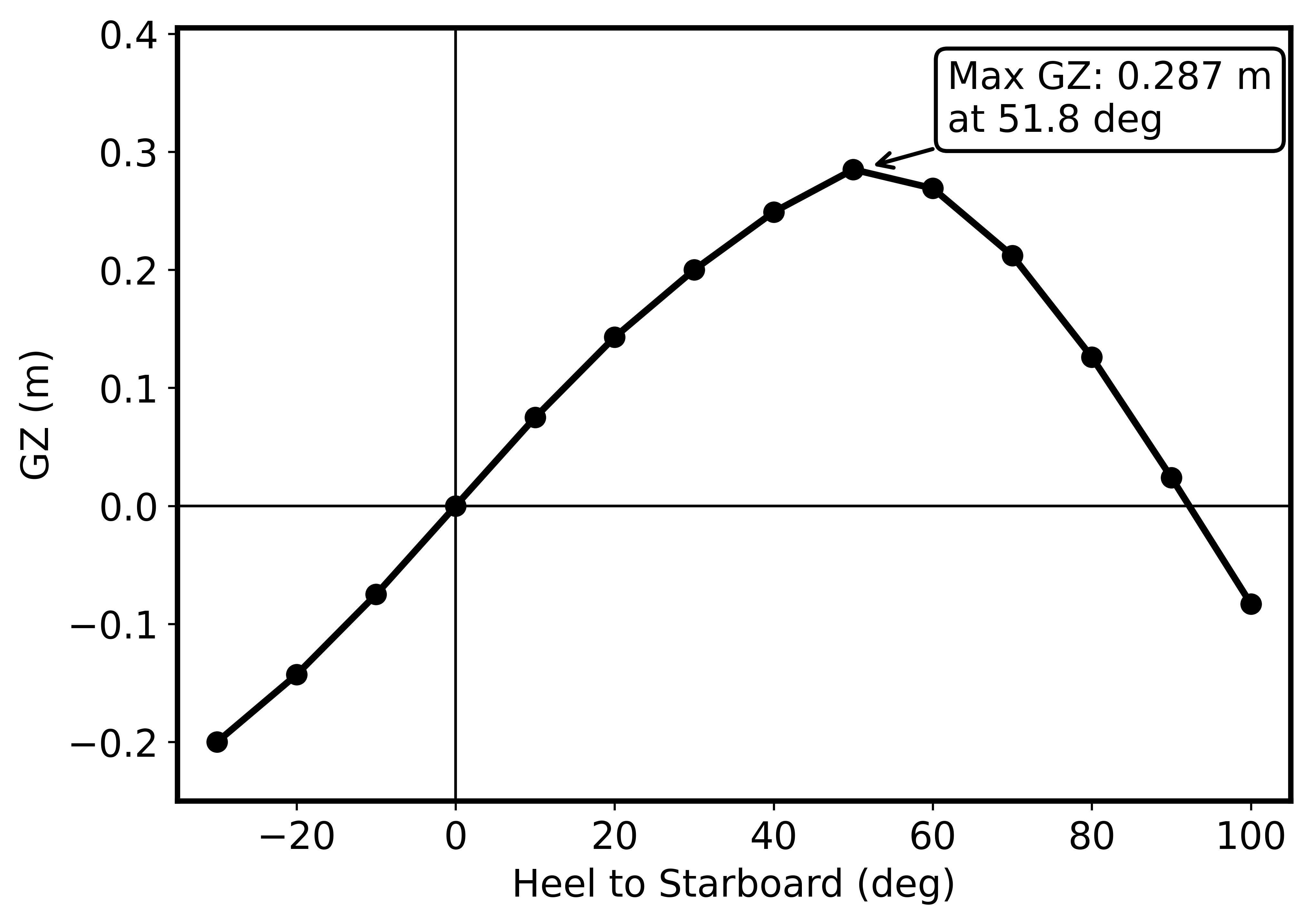}
    \caption{Righting arm (GZ) curve}
    \label{fig:GZcurve}
  \end{subfigure}
  \hfill
  \begin{subfigure}[b]{0.45\textwidth}
    \centering
    \includegraphics[width=\textwidth]{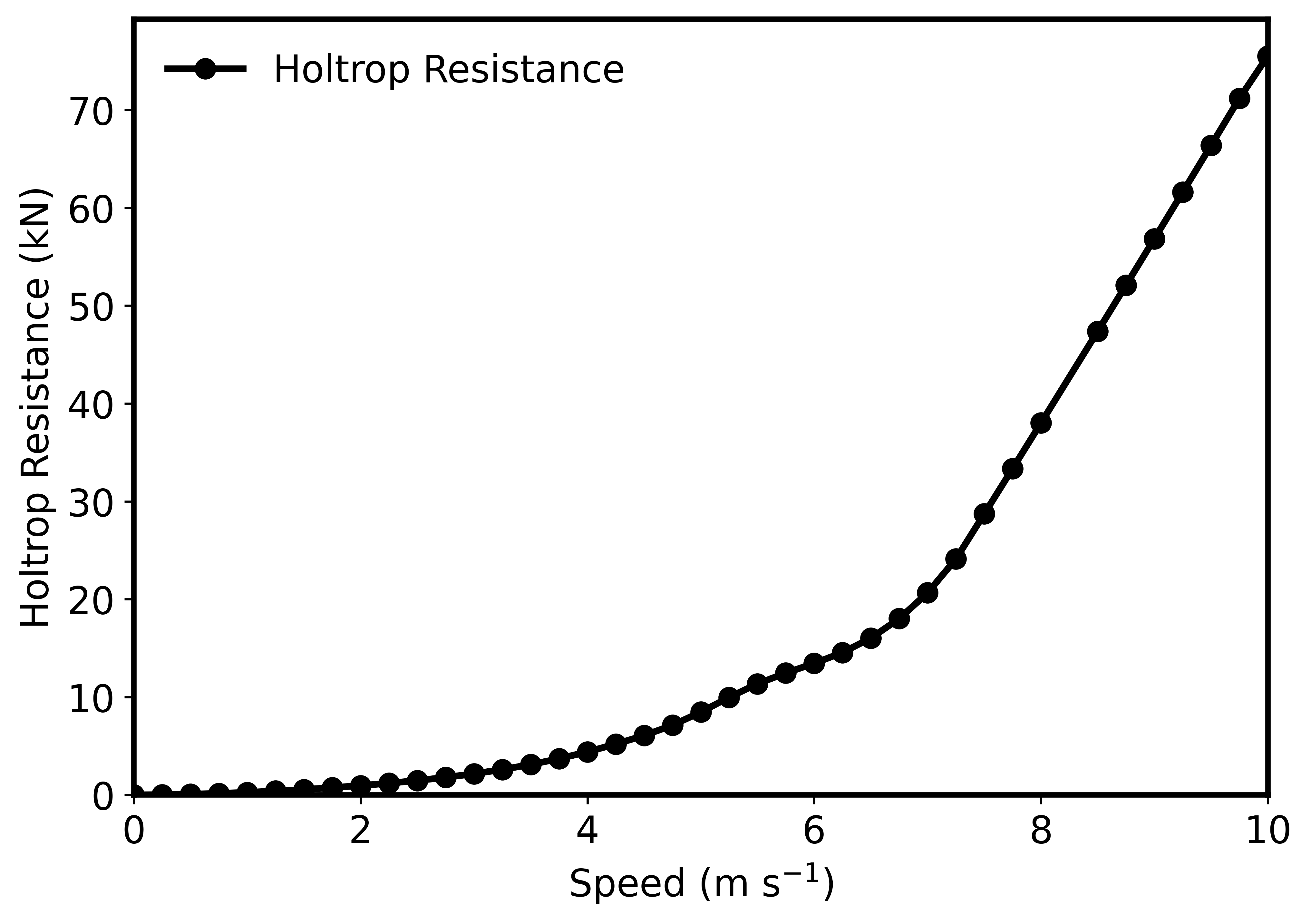}
    \caption{Resistance vs.\ speed curve}
    \label{fig:Resistance}
  \end{subfigure}
  \caption{%
    Validation against \textsc{Maxsurf} simulation: (a)  Righting arm curve for the generated 3D ship hull design, verifying compliance with IMO intact stability criteria, 
    (b) Resistance vs. speed curve confirming smooth hydrodynamic behavior.%
  }
  \label{fig:GZ_and_Resistance}
\end{figure}


Running the hydrostatics routine across a draft sweep (\SI{0.20}{m} to \SI{2.00}{m}) produced the usual particulars table. The design draft adopted for the remainder of the study is \( T = \SI{1.20}{m} \), for which \textsc{Maxsurf} reports a displacement of \( \SI{95.16}{t} \), the longitudinal center of buoyancy \( \mathrm{LCB} = \SI{17.928}{m} \) forward of the aft perpendicular, and the vertical center of buoyancy \( \mathrm{VCB} = \SI{0.697}{m} \) above the baseline. 
Using the same vertical center of gravity, \( KG \), and the design draft, a large angle stability analysis was performed from \(-30^\circ\) to \(100^\circ\) heel. The resulting right-hand curve (GZ) is shown in Figure~\ref{fig:GZcurve}. 
The four quantities demonstrate the suitability of designs, namely (i) linear slope at the origin: \( \frac{dGZ}{d\phi} = 0.434~\text{rad}^{-1} \), confirming a positive initial metacentric height, \( GM > 0 \). (ii) maximum righting arm \( GZ_{\max} = \SI{0.285}{m} \) at a heel angle of \(50^\circ\). (iii) positive range from \(0^\circ\) to \(100^\circ\), indicating that there is no loss of stability before the immersion of the deck edge. (iv) dynamic stability whereby the area under the GZ curve up to \(30^\circ\) heel is \( \SI{0.056}{m \cdot rad} \), and up to \(40^\circ\) heel is \( \SI{0.095}{m \cdot rad} \).
These results exceed the intact stability criteria of the International Maritime Organization (IMO) Resolution for vessels under \SI{100}{m}, which required \( GM_0 \geq \SI{0.15}{m} \) and the area under the GZ curve to \(30^\circ \geq \SI{0.055}{m \cdot rad} \).  The curve remains symmetric, as the model assumes no asymmetric superstructures or wind moments. \textsc{Maxsurf} employs a pure hydrostatic algorithm, recalculating the buoyant center at each angle of the heel and geometrically determining the right arm as \( GZ = GB \cdot \sin \phi \).

The resistance module \textsc{Maxsurf} was set to the Holtrop--Mennen empirical method, which is valid for conventional displacement hulls with $0 < F_r < 0.5 \quad \text{and} \quad 0.15 < C_p < 0.85$.
Our form coefficient \( C_p = 0.706 \) falls well within these limits. For speeds ranging from 0 to 10 $\mathrm{m/s}$, Fig.~\ref{fig:Resistance} shows the curve of total resistance \( R_t \). 
The resulting progressive, hump-free resistance curve is consistent with the relatively fine prismatic coefficient of the learned geometry.

The results for both 2D airfoil and 3D ship hull design tasks demonstrate the effectiveness of the reward-directed diffusion framework in generating high-performance designs without requiring reward gradients. By combining parametric representations with soft value-guided sampling, the model successfully shifts the design distribution beyond the training while preserving physical feasibility. These findings highlight the potential of this approach as a general-purpose tool for data-driven, simulation-based design optimization in computational mechanics.


\section{Conclusion}
\label{sec:conc}
In this study, we proposed a novel generative design framework that incorporates reward guidance during both the training phase and the sampling phases of a diffusion model to generate engineering designs with high performance. By combining fine‑tuning via reward‑weighted maximum likelihood estimation with reward‑based importance sampling during inference, the proposed method achieves significant success in generating high-performance designs. We apply the framework to 2D airfoil and 3D ship hull design tasks and demonstrate its ability to generate high‑performance designs that extend beyond the training data distribution. 

Our empirical results show that the iterative reward-directed framework achieves a reduction of more than 25\% in total resistance for ship hull designs and an improvement of over 10\% in \(C_l/C_d\) for airfoil designs. These design improvements are significant, given that the training datasets, especially in the airfoil case, already reside in optimized design spaces. These results demonstrate the ability of the proposed framework to generalize and extrapolate high‑performing solutions through fine‑tuning that distills soft‑value guidance. By framing the diffusion probabilistic model as a Markov Decision Process and applying iterative directional guidance through soft-value functions, the proposed framework provides a scalable, memory‑efficient optimization pipeline. By limiting the number of candidates per reverse diffusion step to 10 during sampling and leveraging training phase improvements, the framework guides the model towards further optimized designs under memory constraints.

The framework supports integration with simplified and high-fidelity simulation environments and can adapt to various surrogate models, including non‑differentiable architectures. The ability of the pipeline to function without requiring differentiable restrictions enables it to tackle complex design challenges in various applications, as demonstrated in naval architecture and aerodynamic engineering, areas where conventional optimization methods often struggle. Although these results highlight the potential of reward‑directed diffusion models as a powerful, gradient-free approach to performance-driven design, more research is needed to develop a computationally efficient and reliable framework.


\section*{Acknowledgments}
The present study is supported by NSERC Alliance Grant with Elomatic and InnovMarine. This research was also supported in part through computational resources and services provided by Advanced Research Computing (ARC) at the University
of British Columbia and Compute Canada.

\bibliographystyle{elsarticle-num} 
\bibliography{cas-refs}






\end{document}